\pdfoutput=1

\documentclass[11pt]{article}

\usepackage{EMNLP2023}

\usepackage{times}
\usepackage{latexsym}

\usepackage[T1]{fontenc}

\usepackage[utf8]{inputenc}

\usepackage{microtype}

\usepackage{inconsolata}

\usepackage{graphicx}
\usepackage{amsmath}
\usepackage{amsfonts}
\usepackage{booktabs}
\usepackage{threeparttable}
\usepackage{mathrsfs}
\usepackage{latexsym}
\usepackage{stmaryrd}
\usepackage{multicol}
\usepackage{multirow}
\usepackage{xcolor}
\usepackage{algorithm}
\usepackage{algorithmicx}
\usepackage{algpseudocode}

\usepackage{cleveref}
\crefname{section}{§}{§§}
\Crefname{section}{§}{§§}
\usepackage{bbding}
\usepackage{makecell}
\usepackage{url}
\usepackage{arydshln}
\usepackage{subcaption}
\definecolor{sqlcolor}{RGB}{251, 231, 163}
\definecolor{fromcolor}{RGB}{194, 214, 236}
\definecolor{selectcolor}{RGB}{241, 205, 177}
\definecolor{BrickRed}{RGB}{178,34,34}
\title{ASTormer: An AST Structure-aware Transformer Decoder for Text-to-SQL}

\author{Ruisheng Cao, Hanchong Zhang, Hongshen Xu, Jieyu Li, Da Ma, Lu Chen and Kai Yu\thanks{\ \ The corresponding authors are Lu Chen and Kai Yu.}\\
  X-LANCE Lab, Department of Computer Science and Engineering\\
  MoE Key Lab of Artificial Intelligence, AI Institute, Shanghai Jiao Tong University\\
  Shanghai Jiao Tong University, Shanghai China\\
  {\tt \{211314,chenlusz,kai.yu\}@sjtu.edu.cn}\\}

\begin{document}

\maketitle

\begin{abstract}
Text-to-SQL aims to generate an executable SQL program given the user utterance and the corresponding database schema. To ensure the well-formedness of output SQLs, one prominent approach adopts a grammar-based recurrent decoder to produce the equivalent SQL abstract syntax tree~(AST). However, previous methods mainly utilize an RNN-series decoder, which 1) is time-consuming and inefficient and 2) introduces very few structure priors. In this work, we propose an \textbf{AST} structure-aware Transf\textbf{ormer} decoder (ASTormer) to replace traditional RNN cells. The structural knowledge, such as node types and positions in the tree, is seamlessly incorporated into the decoder via both absolute and relative position embeddings. Besides, the proposed framework is compatible with different traversing orders even considering adaptive node selection. Extensive experiments on five text-to-SQL benchmarks demonstrate the effectiveness and efficiency of our structured decoder compared to competitive baselines.
\end{abstract}
\section{Introduction}
\label{sec:intro}

\begin{figure}[t]
    \centering
    \includegraphics[width=0.48\textwidth]{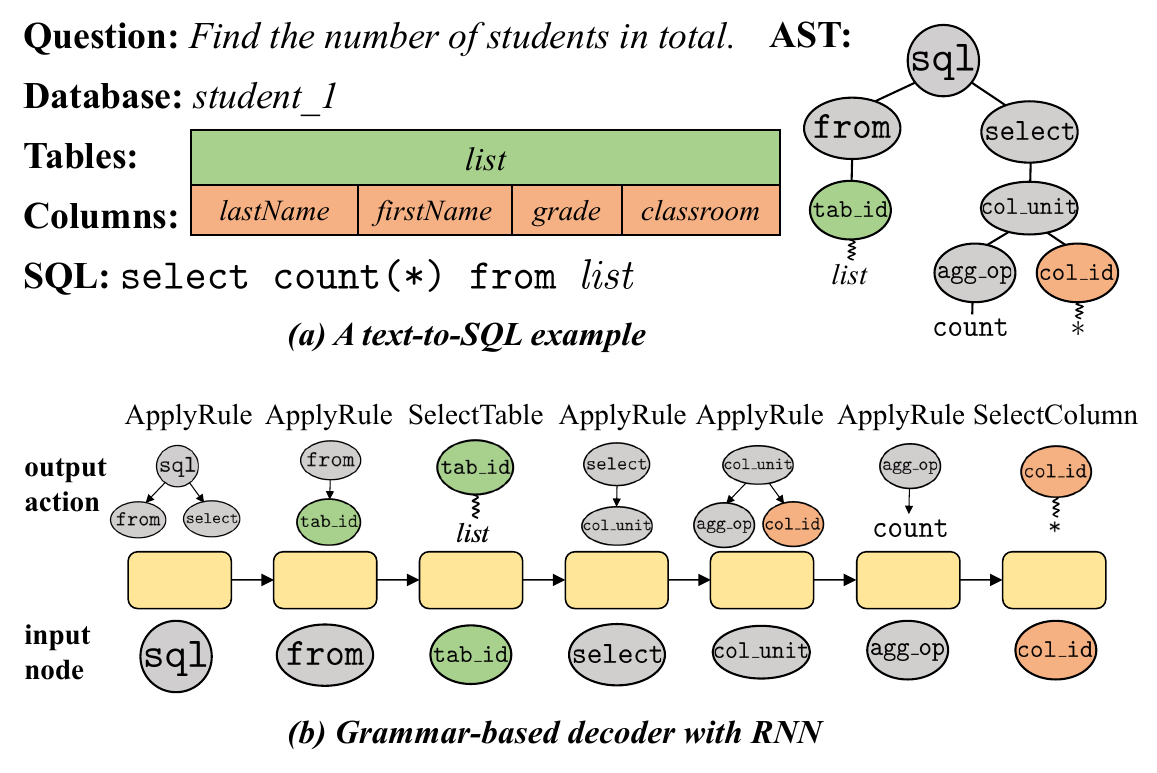}
    \caption{An example of grammar-based text-to-SQL.}
    \label{fig:intro}
\end{figure}

Text-to-SQL~\cite{wikisql, spider} is the task of converting a natural language question into the executable SQL program given the database schema. The mainstream of text-to-SQL parsers can be classified into two categories: token-based sequence-to-sequence parsers~\cite{bridge,picard,unifiedskg} and grammar-based structured parsers~\cite{irnet,ratsql,lgesql}. Token-based parsers treat each token~(including SQL keywords \textsc{Select} and \textsc{From}) in the SQL query as traditional words~(or sub-words). Grammar-based parsers firstly generate the equivalent abstract syntax tree~(AST) of the raw SQL instead, see Figure \ref{fig:intro}(a), and transform the AST into the desired SQL via post-processing. In this work, we focus on the grammar-based branch, because it is more salient in capturing the inner structure of SQL programs compared to purely Seq2Seq methods.

To deal with the structured tree prediction, previous literature mainly utilizes classic LSTM network~\cite{lstm}
to construct the AST step-by-step. This common practice suffers from the following drawbacks: 1) unable to parallelize during training, and 2) incapable of modeling the complicated structure of AST. At each decoding timestep, the auto-regressive decoder chooses one unexpanded node in the partially-generated AST and expands it via restricted actions. Essentially, the process of decoding is a top-down traversal over all AST nodes~(detailed in \cref{sec:astormer}). However, RNN is not specifically designed for structured tree generation and cannot capture the intrinsic connections. To compensate for the structural bias in the recurrent cell, \textsc{TranX}~\cite{tranx} proposes parent feeding strategy, namely concatenating features of the parent node to the current decoder input. Although it achieves stable performance gain, other effective relationships such as siblings and ancestor-descendant are ignored. Besides, most previous work constructs the target AST in a canonical depth-first-search~(DFS) left-to-right~(L2R) order. In other words, the output action at each decoding timestep is pre-defined and fixed throughout training. It potentially introduces unreasonable permutation biases while traversing the structured AST and leads to the over-fitting problem with respect to generation order.

To this end, we propose an \textbf{AST} structure-aware Transf\textbf{ormer} decoder~(ASTormer), and introduce more flexible decoding orders for text-to-SQL. By incorporating features of node types and positions into the Transformer decoder through both absolute and relative position embeddings~\cite{rpr}, dedicated structure knowledge of the typed AST is integrated into the auto-regressive decoder ingeniously. It is also the process of symbol-to-neuron transduction. After computation in the neural space, the probabilistic output action is applied to the SQL AST. New nodes will be attached to the tree, which gives us updated structural information in the symbolic space.
We experiment on five text-to-SQL benchmarks to validate our method, namely Spider~\cite{spider}, SParC~\cite{sparc}, CoSQL~\cite{cosql}, DuSQL~\cite{dusql} and Chase~\cite{chase}. Performances demonstrate the effectiveness and efficiency of ASTormer compared to grammar-based LSTM-series decoders.
To further verify its compatibility with different expanding orders of AST nodes, we try multiple decoding priors, e.g., DFS/breadth-first-search~(BFS) and L2R/random selection. Experimental results prove that the proposed framework is insensitive towards different traversing sequences.

Main contributions are summarized as follows:
\begin{itemize}
    \item A neural-symbolic ASTormer decoder is proposed which adapts the Transformer decoder to structured typed tree generation.
    \item We demonstrate that ASTormer is compatible with different traversing orders beyond depth-first-search and left-to-right~(DFS+L2R).
    \item Experiments on multiple datasets verify the superiority of ASTormer over both grammar- and token-based models on the same scale.
\end{itemize}
\section{Preliminaries}
\label{sec:background}

\subsection{Problem Definition and Overview}
Given a question $Q=(q_1,q_2,\cdots,q_{|Q|})$ with length $|Q|$ and the corresponding database schema $T\cup C$, the goal is to generate the SQL program $y$. The database schema contains multiple tables $T=\{t_i\}_{i=1}^{|T|}$ and columns $C=\{c_i\}_{i=1}^{|C|}$.

A grammar-based text-to-SQL parser follows the generic encoder-decoder~\cite{enc2dec} architecture. The encoder converts all inputs into encoded states $\mathbf{X}=[\mathbf{Q};\mathbf{T};\mathbf{C}]\in\mathbb{R}^{(|Q|+|T|+|C|)\times d}$.
Since it is not the main focus in this work, we assume $\mathbf{X}$ are obtained from a graph-based encoder, such as RATSQL~\cite{ratsql}. 

Next, the decoder constructs the abstract syntax tree~(AST) $y^a$ of SQL $y$ step-by-step via a sequence of actions $\boldsymbol{a}=(a_1,a_2,\cdots,a_{|a|})$. Each action $a_j$ chooses one unexpanded node in the incomplete tree $y^a_{j-1}$ and extends it through pre-defined semantics. The input of the decoder at each timestep is a typed node to expand, while the output is the corresponding action indicating how, see Figure \ref{fig:intro}(b). After the symbolic AST $y^a$ is completed, it is deterministically translated into SQL program $y$.

\subsection{Basic Network Modules}
\begin{figure*}[htbp]
    \centering
    \includegraphics[width=0.95\textwidth]{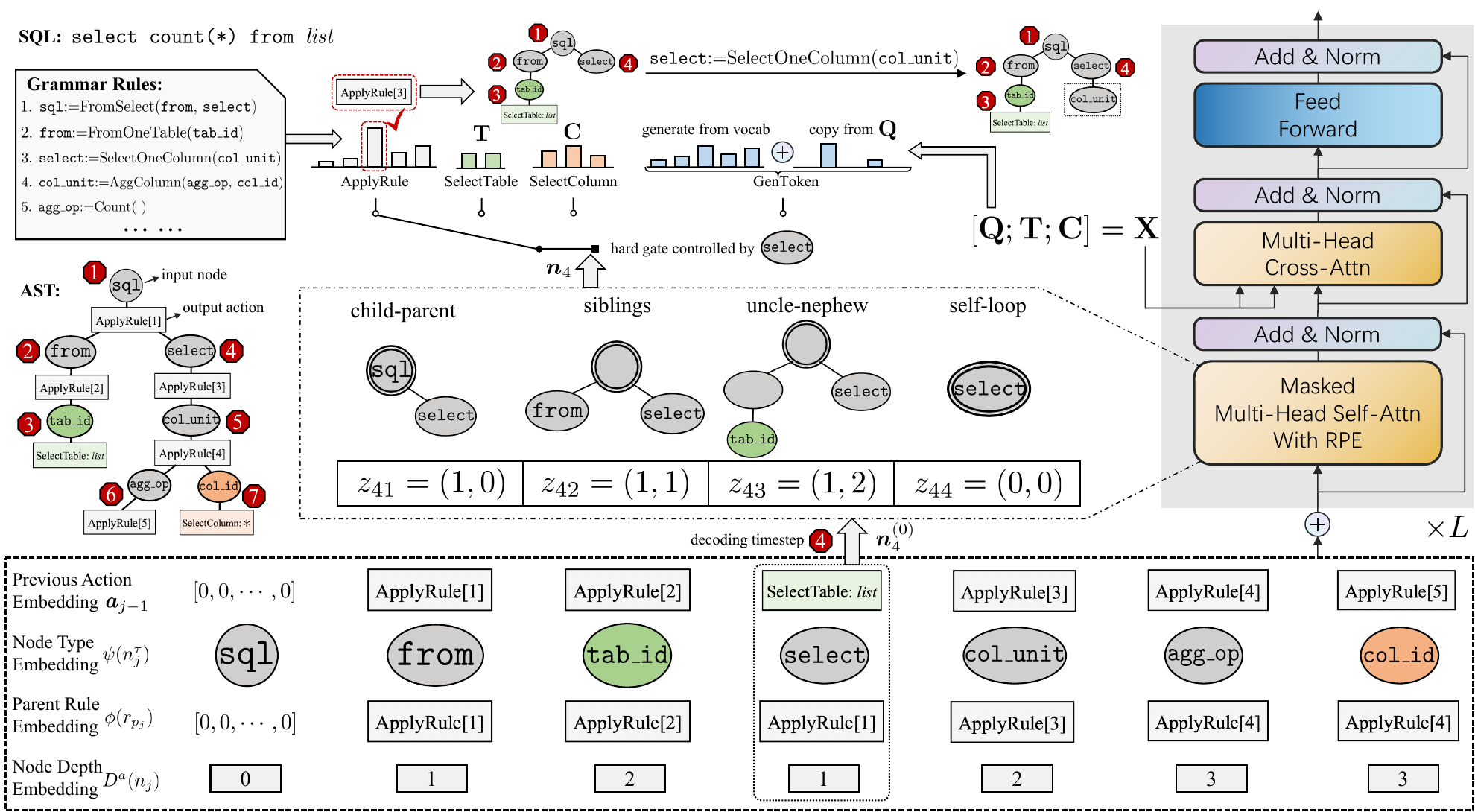}
    \caption{Illustration of ASTormer~(the traversing order is DFS+L2R). We take the decoding timestep $j=4$ as an example. The lowest common ancestor for each node pair is highlighted with double border lines in the middle part.}
    \label{fig:astormer}
\end{figure*}

\paragraph{Transformer Decoder Layer with Relative Position Embedding} Each ASTormer layer consists of $3$ cascaded sub-modules: 1) masked multi-head self-attention with relative position embedding~(RPE), 2) multi-head cross attention, and 3) a feed-forward layer~\cite{transformer}, illustrated in the top right part of Figure~\ref{fig:astormer}. To incorporate target-side self-attention relations $\boldsymbol{z}_{ji}$ between each (query, key) pair $(\boldsymbol{n}_j, \boldsymbol{n}_i)$, we revise the computation of attention weight $\alpha_{ji}^h$ and context vector $\boldsymbol{\tilde{n}}_j^h$ like RPE~\cite{rpr}:~($h$ is head index)
\begin{align*}
\alpha_{ji}^{h}&=\underset{1\le i\le j}{\text{softmax}}\{\frac{(\boldsymbol{n}_j\mathbf{W}_Q^h)(\boldsymbol{n}_i\mathbf{W}_{K}^h+\boldsymbol{z}_{ji})^{\mathrm{T}}}{\sqrt{d/H}}\},\\
\boldsymbol{\tilde{n}}^{h}_j&=\sum_{1\le i\le j}\alpha^h_{ji}(\boldsymbol{n}_i\mathbf{W}^h_V+\boldsymbol{z}_{ji}),
\end{align*}

\noindent Let $\mathbf{N}_{\le j}=[\boldsymbol{n}_1;\cdots;\boldsymbol{n}_j]$ denote the decoder input matrix, $\mathbf{X}=[\boldsymbol{x}_1;\cdots;\boldsymbol{x}_{|X|}]$ denote the encoder cross-attention memory matrix and $\mathbf{Z}_j=\{\boldsymbol{z}_{ji}\}_{i=1}^j$ denote the set of relation features $\boldsymbol{z}_{ji}$ with timestep $i\le j$, we formulate one ASTormer layer as:
\begin{align}
\boldsymbol{n}_j^{\prime}=&\text{ASTormerLayer}(\boldsymbol{n}_j, \mathbf{N}_{\le j}, \mathbf{Z}_j, \mathbf{X}).\label{eq:dec}
\end{align}

\paragraph{Pointer Network} It is widely used to copy raw tokens from the input memory~\cite{see-etal-2017-get}. We re-use the multi-head cross-attention and take the average of weights $\alpha_{ji}^h$ from different heads $h$ as the probability of choosing the $i$-th entry in $\mathbf{X}$:
\begin{align*}
p_{\mathrm{ptr}}(i)=&\frac{1}{H}\sum_{h}a_{ji}^h,\quad h=1,\cdots,H,\\
=&\text{PointerNetwork}(\boldsymbol{n}_j,\mathbf{X})[i].
\end{align*}
\section{ASTormer Decoder}
\label{sec:astormer}
This section introduces our neural-symbolic ASTormer decoder which constructs the SQL AST $y^a$ given encodings $\mathbf{X}=[\mathbf{Q};\mathbf{T};\mathbf{C}]$. The framework is adapted from \textsc{TranX}~\cite{tranx}.

\subsection{Model Overview}
The ASTormer decoder contains three modules: a frontier node input module~(\cref{sec:ape}), a decoder hidden module of $L$ stacked ASTormer layers~(\cref{sec:rpe}), and an action output module~(\cref{sec:prob}). At each decoding timestep $j$, it sequentially performs the following tasks to update the previous AST $y^a_{j-1}$.
\begin{enumerate}
    \item[a).]Select one node $n_j$ to expand~(called \emph{frontier node}) from the frontier node set of $y^a_{j-1}$, and construct input feature $\boldsymbol{n}_j^{(0)}$ for it.
    \item[b).]Further process node $n_j$ in the \emph{neural space} by the decoder hidden module and get the final decoder state $\boldsymbol{n}_j$.
    \item[c).] The action output module computes the distribution $P(a_j)$ based on the type of node $n_j$.
    \item[d).]Choose a valid action $a_j$ from $P(a_j)$ and update $y^a_{j-1}$ in the \emph{symbolic space}.
\end{enumerate}
These steps are performed recurrently until no frontier node can be found in $y^a_j$. In other words, all nodes in $y^a_j$ have been expanded.

\subsection{Basic Concepts of AST}
To recap briefly, an abstract syntax tree~(AST) is a tree representation of the source code.
\label{sec:node_type}
\paragraph{Node type} Each node $n$ in the AST is assigned a \emph{type} attribute $n^{\tau}$ representing its syntactic role. For example, in Figure~\ref{fig:astormer}, one node with type {\tt sql} indicates the root of a complete SQL program.
Nodes can be classified into non-terminal and terminal nodes based on their types.
In this task, terminal types include {\tt tab\_id}, {\tt col\_id}, and {\tt tok\_id}, which denote the index of table, column, and tokens, respectively. Embeddings of each node type $n^{\tau}$ is provided by function $\psi(n^{\tau})\in\mathbb{R}^{1\times d}$.

\paragraph{Grammar rule} Each rule $r$ takes the form of
\begin{center}
{\tt p$\_$type} := RuleName({\tt c$\_$type1, c$\_$type2,$\cdots$}),
\end{center}
where {\tt p$\_$type} is the type of the parent non-terminal node to expand, while ({\tt c$\_$type1, $\cdots$}) denote types of children nodes to attach~(can be duplicated or empty). Only non-terminal types can appear on the left-hand side of one rule. The complete set of grammar rules are list in Appendix \ref{app:grammar}. We directly retrieve the feature of each grammar rule $r$ from an embedding function $\phi(r)\in\mathbb{R}^{1\times d}$.

\subsection{Frontier Node Input Module}
\label{sec:ape}
Given the chosen frontier node $n_j$, we need to firstly construct the input feature $\boldsymbol{n}_j^{(0)}$ at timestep $j$. It should reflect the information of node type and position in the AST to some extent. Thus, we initialize it as a sum of four vectors, including:
\begin{enumerate}
    \item[1.] Previous action embedding $\boldsymbol{a}_{j-1}$~(defined in \cref{sec:prob}), to inform the decoder of how to update the incomplete AST in the neural space.
    \item[2.] Node type embedding $\psi(n^{\tau}_j)$ of frontier node.
    \item[3.] Parent rule embedding $\phi(r_{p_j})$, where $p_j$ denotes the timestep when the parent of node $n_j$ is expanded. This is similar to the parent feeding strategy in \citet{syntactic}.
    \item[4.] Depth embedding $D^a(n_j)\in\mathbb{R}^{1\times d}$ which indicates the depth of node $n_j$ in the AST $y^a_{j-1}$.
\end{enumerate}
Notice that, vectors 2-4 serve as the original absolute position embeddings~(APE) in Transformer decoder. We apply a LayerNorm~\cite{layernorm} layer upon the sum of these vectors. Formally,
$$\boldsymbol{n}_j^{(0)}=\text{LN}(\boldsymbol{a}_{j-1}+\psi(n^{\tau}_j)+\phi(r_{p_j})+D^a(n_j)).$$

\subsection{Decoder Hidden Module}
\label{sec:rpe}
The initial node feature $\boldsymbol{n}_j^{(0)}$ is further processed by $L$ stacked ASTormer layers, see Eq.(\ref{eq:dec}). The final decoder hidden state $\boldsymbol{n}_j$ is computed by
\begin{align*}
\boldsymbol{n}_j^{(l+1)}&=\text{ASTormerLayer}^{(l)}(\boldsymbol{n}_j^{(l)}, \mathbf{N}^{(l)}_{\le j}, \mathbf{Z}_j,\mathbf{X}),\\
\boldsymbol{n}_j&=\boldsymbol{n}_j^{(L)},\quad l=0,\cdots,L-1,
\end{align*}
where $\mathbf{N}^{(l)}_{\le j}$ denotes all node features $\boldsymbol{n}_i^{(l)}$ with timestep $i\le j$ from the $l$-th decoder layer, and $\mathbf{Z}_j$ denotes all relational features $\boldsymbol{z}_{ji}$ from node $n_j$ to each previous node $n_i$.
Borrowing the concept of \emph{lowest common ancestor}~(LCA) in data structure~\cite{lca}, the relation $z_{ji}$ between each node pair $(n_j,n_i)$ is defined as a tuple:
\begin{align}
\text{LCA}(n_k,n_s)&=\text{the LCA of node }n_k\text{ and }n_s,\notag\\
\text{dist}(n_k,n_s)&=\text{the distance between }n_k\text{ and }n_s,\notag\\
p_j&=\text{dist}(\text{LCA}(n_j, n_i),n_j),\notag\\
p_i&=\text{dist}(\text{LCA}(n_j, n_i), n_i),\notag\\
z_{ji}=&(\text{clamp}(p_j,R), \text{clamp}(p_i,R)),\label{eq:rdist}
\end{align}
where $\text{clamp}(\cdot,R)$ truncates the maximum distance to $R$. Specifically, $z_{ji}=(1,0)$ implies that $n_j$ is a child of node $n_i$, while $z_{ji}=(1,1)$ denotes sibling relationships. By definition, $z_{ji}$ is symmetric in that if $z_{ji}=(2,3)$, the reverse relation must be $z_{ij}=(3,2)$. However, notice that some relations such as $z_{ji}=(0,1)$ will never be used, because a descendant node will not be expanded until all ancestors are done during the top-down traversal of AST. This topological constraint is fulfilled by the triangular future mask in self-attention.

Combining APE and RPE, the entire AST can be losslessly reconstructed through node types of endpoints and relative relations between any node pairs. For training, these information can be pre-extracted once and for all. During inference, the relation $Z_j$ can be efficiently constructed on-the-fly via dynamic programming due to the symmetric definition of $z_{ji}$ and acyclic property of AST~(detailed in Appendix~\ref{app:on-the-fly}).

\subsection{Action Output Module}
\label{sec:prob}
\begin{figure}[htbp]
    \centering
    \includegraphics[width=0.45\textwidth]{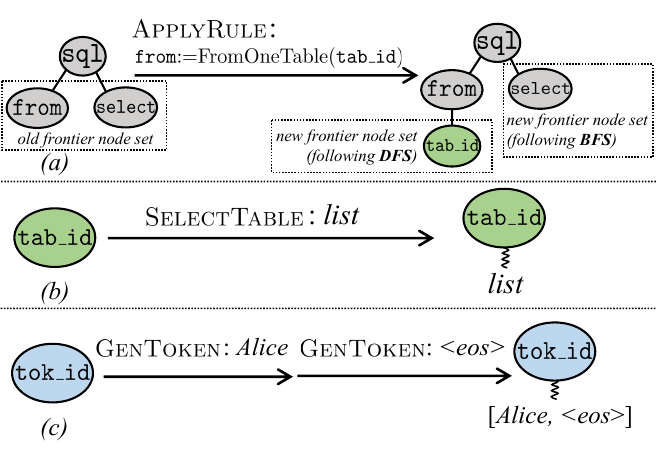}
    \caption{Three types of actions and their semantics.}
    \label{fig:action}
\end{figure}

Given decoder state $\boldsymbol{n}_j$, the next step is to compute the distribution $P(a_j)$ of output actions based on the node type $n^{\tau}_j$. As illustrated in Figure \ref{fig:action}, there are three genres of actions: 1) \textsc{ApplyRule}$[r]$ for non-terminal types, 2) \textsc{SelectItem}$[i]$ for terminal types {\tt tab\_id} and {\tt col\_id}~(\textsc{Item}$\in$$\{$\textsc{Table}, \textsc{Column}$\}$), and 3) \textsc{GenToken} for {\tt tok\_id}.
\paragraph{\textsc{ApplyRule} Action} This action chooses one rule $r_i$ from a constrained set of grammar rules determined by the non-terminal type $n^{\tau}_j$. Specifically, the left-hand side of grammar rule $r_i$ must equal $n^{\tau}_j$~(known as node type constraint,~\citealp{type-constraints}). When applying this action to AST $y_{j-1}^a$ in the symbolic space, we attach all children types on the right-hand side of rule $r_i$ to the parent node $n_j$ and mark these children nodes as unexpanded. Given $\boldsymbol{n}_j$, the distribution of \textsc{ApplyRule} action is calculated by
\begin{multline*}
P(a_{j}=\textsc{ApplyRule}[r_i]|a_{<j},\boldsymbol{n}_j,\mathbf{X})=\\
\underset{i}{\text{softmax}}\{\boldsymbol{n}_j\mathbf{W}_{\text{ar}}\phi(r_i)^{\textrm{T}}\}.
\end{multline*}
\paragraph{\textsc{SelectItem} Action} Taking \textsc{SelectTable} action as an example, to expand frontier node with terminal type {\tt tab$\_$id}, we directly choose one table entry and attach it to the input terminal node. The probability of selecting the $i$-th entry from encoded table memory $\mathbf{T}\in\mathbb{R}^{|T|\times d}$ is computed by
\begin{multline*}
P(a_{j}=\textsc{SelectTable}[t_i]|a_{<j},\boldsymbol{n}_j,\mathbf{X})=\\
\text{PointerNetwork}_{\text{st}}(\boldsymbol{n}_j,\mathbf{T})[i].
\end{multline*}
\paragraph{\textsc{GenToken} Action} To produce SQL values for terminal node of type {\tt tok\_id}, we utilize the classic pointer-generator network~\cite{see-etal-2017-get}. Raw token $w_i$ can be generated from a vocabulary or copied from the input question memory $\mathbf{Q}\in\mathbb{R}^{|Q|\times d}$. Mathematically,
\begin{align*}
g_j&=\text{sigmoid}([\boldsymbol{n}_j\circ\boldsymbol{n}_j^{(0)}]\mathbf{W}_{\text{g}}),\\
P_{\text{gen}}(w_i)&=\underset{i}{\text{softmax}}\{\boldsymbol{n}_j\mathbf{W}_{\text{gen}}e(w_i)^{\mathrm{T}}\},\\
P_{\text{copy}}(w_i)&=\sum_{k:\ q_k=w_i}\text{PointerNetwork}_{\text{gt}}(\boldsymbol{n}_j, \mathbf{Q})[k],\\
\begin{split}
P(a_j&=\textsc{GenToken}[w_i]|a_{<j},\boldsymbol{n}_j,\mathbf{X})=\\
    &\quad g_j P_{\text{gen}}(w_i)+ (1-g_j)P_{\text{copy}}(w_i),
\end{split}
\end{align*}
where $g_j\in[0,1]$ is the balance score and $e(w_i)\in\mathbb{R}^{1\times d}$ denotes the word embedding of token $w_i$. Since each SQL value may have multiple tokens, node $n_j$ is  not expanded until a special ``<eos>" token is emitted. Intermediate results are buffered and the frontier node chosen at timestep $j+1$ is still $n_j$ if the current output action $a_j$ is not ``<eos>". After the emission of ``<eos>", we extend node $n_j$ by attaching the list of all buffered tokens.

Until now, the action embedding $\boldsymbol{a}_j\in\mathbb{R}^{1\times d}$~(as decoder input for the next timestep) can be defined,
\begin{align*}
\boldsymbol{a}_j=\begin{cases}
\phi(r_j) & \text{if } a_j\in\textsc{ApplyRule}\\
\boldsymbol{t}_j \text{ or } \boldsymbol{c}_j& \text{if }a_j\in\textsc{SelectItem}\\
e(w_j) & \text{if }a_j\in\textsc{GenToken}
\end{cases},
\end{align*}
where $r_j$/$w_j$ is the rule/token chosen at the current timestep $j$, and $\boldsymbol{t}_j$/$\boldsymbol{c}_j$ is the selected entry in table/column memory $\mathbf{T}$/$\mathbf{C}$. Parameters of $\text{PointerNetwork}_{\ast}(\boldsymbol{n}_j,\cdot)$ are shared among \textsc{SelectTable}, \textsc{SelectColumn}, and \textsc{GenToken} actions to avoid over-parametrization.

The prediction of the SQL AST $y^a$ can be decoupled into a sequence of actions $\boldsymbol{a}=(a_1,\cdots,a_{|a|})$. The training objective for the text-to-SQL task is 
\begin{align}
\mathcal{L}=-\sum_{j=1}^{|a|} \log P(a_j|a_{<j},\mathbf{X}).\label{eq:loss}
\end{align}

\subsection{Different Traversal Paths}
\label{sec:order}
What remains to be resolved is 1) how to update the frontier node set given a new action $a_j$, and 2) choose one input frontier node $n_j$ from this set.
\subsubsection{Update Frontier Node Set: DFS or BFS}
The frontier node set refers to a restricted set of unexpanded nodes in the incomplete AST $y^a_{j-1}$ that we focus on currently. It is usually constrained and implemented by depth-first-search~(DFS) or breadth-first-search~(BFS).
If DFS is strictly followed, the sets of unexpanded nodes can be stored as a \emph{stack}. Each element in the stack is a set of unexpanded nodes. After applying action $a_j$ to frontier node $n_j$, we remove $n_j$ from the node set at the top of stack and push the set of its children nodes~(if not empty) onto the stack. By analogy, if the AST traversal order is breadth-first-search~(BFS), we only need to maintain a first-in-first-out \emph{queue}. For example, in Figure~\ref{fig:action}(a), after expanding node ``{\tt from}'', the updated frontier node set is the singleton set which contains node ``{\tt tab\_id}'' if DFS is followed~(otherwise node ``{\tt select}'' if BFS). Note that, if $n_j^{\tau}={\tt tok\_id}$ and the current output action $a_j\neq\textsc{GenToken}[\text{<eos>}]$, the frontier node $n_j$ is preserved and forced to be chosen next.

\subsubsection{Choose Frontier Node: L2R or Random}
To select frontier node from the restricted set, a common practice~\cite{irnet} is to stipulate a canonical order, also known as left-to-right~(L2R). For example, given the grammar rule {\tt sql} := FromSelect({\tt from}, {\tt select}), the model always expands node ``{\tt from}'' prior to ``{\tt select}". Evidently, this pre-defined priority abridges the freedom in AST construction. In this work, we attempt to eliminate this constraint and propose a more intuitive Random method. 
Concretely, during training, we randomly sample one frontier node from the node set at top of stack~(or head of queue). Such randomness prevents the model from over-fitting the permutation bias and inspires the decoder to truly comprehend the relative positions $z_{ji}$ of different node pairs. As for inference, instead of choosing one candidate from the frontier node set at each timestep, all distinct options are enumerated as inputs to the decoder to enlarge the search space. This relaxation encourages the model itself to discover the optimal traversal path and provides more interpretation on the neural-symbolic processing.

Due to page limit, more details and the complete algorithm are elaborated in Appendix \ref{app:alg}.
\section{Experiments}
\label{sec:exp}
\begin{table*}[htbp]
  \centering
  \resizebox{0.95\textwidth}{!}{
    \begin{tabular}{c|cc|cc|cc|cc}
    \hline

    \hline
    \multirow{2}{*}{\textbf{Category}} & \multirow{2}{*}{\textbf{Method}} & \multirow{2}{*}{\textbf{PLM}} & \multicolumn{2}{c|}{\textbf{Spider}} & \multicolumn{2}{c|}{\textbf{SParC}} & \multicolumn{2}{c}{\textbf{CoSQL}} \\
\cline{4-9}          &       &       & \textbf{EM} & \textbf{EX} & \textbf{QM} & \textbf{IM} & \textbf{QM} & \textbf{IM} \\
    \hline\hline
    \multirow{3}{*}{token-based} & EditSQL~\cite{editsql} & \multirow{3}{*}{BERT~\cite{bert}} & 57.6  & -     & 47.2  & 29.5  & 39.9  & 12.3 \\
          & GAZP~\cite{gazp}  &       & 59.1  & 59.2  & 48.9  & 29.7  & 42.0  & 12.3 \\
          & IGSQL~\cite{igsql} &       & -     & -     & 50.7  & 32.5  & 44.1  & 15.8 \\
    \hline
    \multirow{2}{*}{grammar-based} & RATSQL~\cite{ratsql} & \multirow{2}{*}{BERT} & 69.7  & -     & -     & -     & -     & - \\
          & R$^2$SQL~\cite{r2sql} &       & -     & -     & 54.1  & 35.2  & 45.7  & \textbf{19.5} \\
    \hdashline
    \multirow{2}{*}{\textbf{Ours}} & RATSQL w/ LSTM & \multirow{2}{*}{BERT} &  70.1   &    67.2   &  58.7   &  38.2  &  49.1   &  18.1  \\
          & \textbf{RATSQL w/ ASTormer} &       & \textbf{71.7} & \textbf{70.4} & \textbf{61.6} & \textbf{39.8} & \textbf{49.9} & 18.4 \\
    \hline

    \hline
    \multirow{2}{*}{token-based} & \textsc{UnifiedSKG}~\cite{unifiedskg} & T5-Large~\cite{t5} & 67.6  & -     & 59.0  & -     & 51.6  & - \\
          & \textsc{UniASr}~\cite{unisar} & BART~\cite{bart}  & 70.0  & 60.4  & 60.4  & 40.8  & 51.8  & 21.3 \\
    \hline
    \multirow{3}{*}{grammar-based} & RATSQL & \textsc{GraPPa}~\cite{grappa} &    73.6   &  -     & -     & -     & -     & - \\
          & RATSQL & \textsc{SCoRe}~\cite{score} & -     & -     & 62.2  & 42.5  & 52.1  & 22.0  \\
          & HIESQL~\cite{hiesql} & \textsc{GraPPa}~\cite{grappa} & -     & -     & 64.7  & \textbf{45.0}  & \textbf{56.4} & \textbf{28.7} \\
    \hdashline
    \multirow{2}{*}{\textbf{Ours}} & RATSQL w/ LSTM & \multirow{2}{*}{ELECTRA~\cite{electra}} &  72.9 &  71.4 &  62.8  &    41.2   &   52.9  & 22.5 \\
          & \textbf{RATSQL w/ ASTormer} &  & \textbf{74.6}  & \textbf{73.2}  & \textbf{64.8} & \textbf{45.0} & 55.5  & 22.9 \\
    \hline

    \hline
    \end{tabular}%
}
  \caption{Main results on three English benchmarks. For fairness, we compare results with those using PLMs on the same scale. We do not utilize any reranking, regularization, data augmentation, and ensemble methods.}
  \label{tab:main}%
\end{table*}%

\subsection{Experimental Setup}
\paragraph{Datasets} We experiment on $5$ typical text-to-SQL benchmarks. Spider~\cite{spider} is a large-scale, cross-domain, multi-table English benchmark for text-to-SQL. SparC~\cite{sparc} and CoSQL~\cite{cosql} are multi-turn versions. To compare ASTormer with models in the literature, we report official metrics on development sets~(EM and EX accuracy for Spider, QM and IM for SParC and CoSQL), since test sets are not publicly available. For results on Chinese datasets such as DuSQL~\cite{dusql} and Chase~\cite{chase}, see Appendix \ref{app:chinese} for details.

Our model is implemented with Pytorch~\cite{pytorch}~\footnote{\url{https://github.com/rhythmcao/astormer}}. PLM and its initial checkpoint are downloaded from the transformers~\cite{huggingface} library. For the graph encoder, we reproduce the prevalent RATSQL~\cite{ratsql}. The hidden dimension $d$ is $512$ and the number of encoder/decoder layers is $8$/$2$. The number of heads and dropout rate is set to $8$ and $0.2$ respectively. Throughout the experiments, we use AdamW~\cite{adam} optimizer with a linear warmup scheduler. The warmup ratio of total training steps is $0.1$. The leaning rate is $4e\textrm{-}4$~({\tt small}), $2e\textrm{-}4$~({\tt base}) or $1e\textrm{-}4$~({\tt large}) for PLMs of different sizes and the weight decay rate is fixed to $0.1$. The optimization of PLM parameters is carried out more carefully with learning rate layerwise decay~(coefficient $0.8$). Batch size is $20$, and the number of training iterations is $100k$. For inference, we adopt beam search with size $5$.

\subsection{Main Results}
Main results are provided in Table \ref{tab:main}. Some conclusions can be summarized: 1) Our ASTormer decoder consistently outperforms the LSTM decoder across all three benchmarks on both metrics. For example, with ELECTRA~\cite{electra}, ASTormer beats traditional LSTM decoder by $1.7$ points on Spider in metric EM, while $2$ or more points on SParC and CoSQL. Moreover, we compute the average training time per $100$ iterations for LSTM and ASTormer decoders in Table \ref{tab:time}. Not surprisingly, ASTormer is three times faster~(even four times on Spider due to longer action sequences) than LSTM decoder. Although it introduces some overheads about structural features, the overall training time is still comparable to token-based Transformer on account of shorter output sequences. That is, the average number of AST nodes is smaller than that of tokenized SQL queries, see Appendix \ref{app:dataset} for dataset statistics. 2) Arguably, grammar-based decoders are superior to token-based methods among models on the same scale. It can be explained that grammar-based methods explicitly inject the structure knowledge of SQL programs into the decoder. Although recent advanced token-based models such as \textsc{Picard}\cite{picard} and RASAT\cite{rasat} show exciting progress, they heavily rely on large PLMs, which may be unaffordable for some research institutes. Thus, lightweight grammar-based models provide a practical solution in resource-constrained scenarios. 3) Equipped with more powerful PLMs tailored for structured data, performances can be further promoted stably. Accordingly, we speculate those task-adaptive PLMs which focus on enhancing the discriminative capability are more suitable in text-to-SQL. The evidence is that PLMs such as \textsc{GraPPa}~\cite{grappa}, \textsc{SCoRe}~\cite{score}, and ELECTRA~\cite{electra}, all design pre-training tasks like syntactic role classification and word replacement discrimination. Pre-training on large-scale semi-structured tables is still a long-standing and promising direction.
\begin{table}[tp]
  \centering
  \resizebox{0.49\textwidth}{!}{
    \begin{tabular}{c|c|c|c|l}
    \hline

    \hline
    \textbf{Dataset} & \textbf{Category} & \textbf{Cell}  & \textbf{Acc.}  & \multicolumn{1}{c}{\textbf{Time/$sec.$}} \\
    \hline
    \multirow{3}{*}{Spider} & token-based & Transformer & $69.7$ & $49.5$ \\
\cdashline{2-5}[1pt/1pt]        & \multirow{2}{*}{grammar-based} & LSTM  & $69.9$ & $191.4$ \\
          &       & ASTormer & $71.4$ & $45.8${\color{darkblue}\bf\tiny $(\times 4.2)$} \\
    \hline
    \multirow{3}{*}{SParC} & token-based & Transformer & $59.8$ & $49.1$ \\
\cdashline{2-5}[1pt/1pt]           & \multirow{2}{*}{grammar-based} & LSTM &  $60.9$  & $168.1$ \\
          &       & ASTormer &  $61.3$ & $49.8${\color{darkblue}\bf\tiny $(\times 3.4)$} \\
    \hline
    \multirow{3}{*}{CoSQL} & token-based & Transformer & $50.1$ & $52.8$ \\
\cdashline{2-5}[1pt/1pt]          & \multirow{2}{*}{grammar-based} & LSTM  & $50.5$  & $154.2$ \\
          &       & ASTormer & $51.1$ & $60.7${\color{darkblue}\bf\tiny $(\times 2.5)$} \\
    \hline

    \hline
    \end{tabular}%
    }
  \caption{Acc. and avg. training time~(seconds) per $100$ iterations with PLM {\tt electra-small}. We also implement a token-based Transformer decoder with question copy and schema selection mechanism~\cite{see-etal-2017-get}.}
  \label{tab:time}%
\end{table}%

\subsection{Ablation Studies}
For the sake of memory space, the following experiments are conducted with PLM {\tt electra-small} unless otherwise specified.
\begin{table}[tp]
  \centering
    \resizebox{0.49\textwidth}{!}{
    \begin{tabular}{l|lll}
    \hline

    \hline
    \multicolumn{1}{c|}{\textbf{Method}} & \multicolumn{1}{c}{\textbf{Spider}} & \multicolumn{1}{c}{\textbf{SParC}} & \multicolumn{1}{c}{\textbf{CoSQL}} \\
    \hline\hline
    \textbf{ASTormer} & \multicolumn{1}{c}{\textbf{71.4}}  & \multicolumn{1}{c}{\textbf{61.3}}  & \multicolumn{1}{c}{\textbf{51.1}} \\
    \hline
    \quad w/o node type & 70.9{\color{BrickRed}\bf\tiny $(0.5\downarrow)$} & 58.5{\color{BrickRed}\bf\tiny $(2.8\downarrow)$}   & 48.3{\color{BrickRed}\bf\tiny $(2.8\downarrow)$} \\
    \quad w/o parent rule & 69.0{\color{BrickRed}\bf\tiny $(2.4\downarrow)$}  & 58.6{\color{BrickRed}\bf\tiny $(2.7\downarrow)$}  & 49.4{\color{BrickRed}\bf\tiny $(1.7\downarrow)$}  \\
    \hdashline
    \quad w/o node depth & 70.1{\color{BrickRed}\bf\tiny $(1.3\downarrow)$}  & 58.2{\color{BrickRed}\bf\tiny $(3.1\downarrow)$}  & 47.1{\color{BrickRed}\bf\tiny $(4.0\downarrow)$}  \\
    \quad w/ pe & 69.0{\color{BrickRed}\bf\tiny $(2.4\downarrow)$}  & 58.9{\color{BrickRed}\bf\tiny $(2.4\downarrow)$}  & 48.9{\color{BrickRed}\bf\tiny $(2.2\downarrow)$}  \\
    \hdashline
    \quad w/o relations & 69.7{\color{BrickRed}\bf\tiny $(1.7\downarrow)$}  & 58.9{\color{BrickRed}\bf\tiny $(2.4\downarrow)$}  & 48.1{\color{BrickRed}\bf\tiny $(3.0\downarrow)$}  \\
    \quad w/ rpe $j-i$ & 69.4{\color{BrickRed}\bf\tiny $(2.0\downarrow)$} &  58.2{\color{BrickRed}\bf\tiny $(3.1\downarrow)$} &  48.9{\color{BrickRed}\bf\tiny $(2.2\downarrow)$} \\
    \hline
    \textbf{Transformer} & 68.2{\color{BrickRed}\bf\tiny $(3.2\downarrow)$}  & 57.3{\color{BrickRed}\bf\tiny $(4.0\downarrow)$} &  47.5{\color{BrickRed}\bf\tiny $(3.6\downarrow)$} \\
    \hline

    \hline
    \end{tabular}%
    }
  \caption{Ablation of different components in ASTormer. ``pe''/``rpe'' represents traditional absolute and relative position embeddings respectively; ``Transformer'' denotes an AST decoder which only utilizes previous action embedding and position embedding. }
  \label{tab:ablate_pe}%
\end{table}%

\begin{table}[tp]
  \centering
    \resizebox{0.49\textwidth}{!}{
    \begin{tabular}{c|cccc|c}
    \hline
    
    \hline
    \textbf{Size of $R$} & \textbf{2} & \textbf{4} & \textbf{8} & \textbf{16} & \textbf{Max Tree Depth}\\
    \hline\hline
    Spider & 69.9  & 71.3  & \textbf{71.4} & 70.4 & 16 \\
    SParC & 60.8  & \textbf{61.3} & 60.4  & 61.2 & 10 \\
    CoSQL & 49.8  & \textbf{51.1} & 50.3  & 50.0 & 10 \\
    \hline

    \hline
    \end{tabular}%
    }
  \caption{Ablation of the maximum relative distance $R$.}
  \label{tab:ablate_astormer}%
\end{table}%

\paragraph{Ablation of Structural Features} In Table \ref{tab:ablate_pe}, we analyze the contribution of each module in ASTormer, including the node type embedding $\psi(n_j^{\tau})$, the parent rule embedding $\phi(r_{p_j})$, the node depth embedding $D^a(n_j)$, and the relative positions between node pairs $z_{ji}, i\le j$. From the overall results, we can find that the performance undoubtedly declines no matter which component is omitted. When we replace the node depth/relations $z_{ji}$ with traditional absolute/relative position embeddings, the performance still decreases. It demonstrates that traditional position embeddings are not suitable in structured tree generation and may introduce incorrect biases. When we remove all structural features, the proposed ASTormer decoder degenerates into purely Transformer-based AST decoder. Unfortunately, it also gives the worst performances, which validates the significance of integrating structure knowledge into the decoder from the opposite side.
\paragraph{Impact of Number of Relations} Next, we attempt to find the optimal number of relation types. As defined in Eq.(\ref{eq:rdist}), threshold $R$ is used to truncate the maximum distance between node $n_j$ and the lowest common ancestor of $(n_j,n_i)$. Notice that, the number of total relation types increases at a square rate as the threshold $R$ gets larger. The original problem is reduced to find a suitable threshold $R$. Intuitively, the optimal threshold is tightly related to both the shape and size of SQL ASTs. We calculate the maximum tree depth for each dataset in the last column of Table \ref{tab:ablate_astormer}. By varying the size of $R$, we discover that one feasible choice is to set $R$ as roughly half of the maximum tree depth.




\begin{table}[htbp]
  \centering
  \resizebox{0.45\textwidth}{!}{
    \begin{tabular}{l|ccc}
    \hline

    \hline
    \multicolumn{1}{c|}{\textbf{Decoding Order}} & \textbf{Spider} & \textbf{Sparc} & \textbf{CoSQL} \\
    \hline
    DFS+L2R & 68.5  & \textbf{58.6} & 48.7  \\
    DFS+Random & \textbf{68.6} & 58.1  & 48.5  \\
    \hline
    BFS+L2R & 68.2  & 58.2  & \textbf{48.9} \\
    BFS+Random & 68.0  & 57.2  & 47.1  \\
    \hline

    \hline
    \end{tabular}%
  }
  \caption{Ablation of different traversal orders. DFS: depth-first-search; BFS: breadth-first-search; L2R: left-to-right; Random: random sampling during training.}
  \label{tab:ablate_order}%
\end{table}%

\paragraph{Different Traversal Orders}
We also compare $4$ different choices depending on how to update the frontier node set~(DFS or BFS) and how to select frontier node from this set~(L2R or Random), detailed in \cref{sec:order} and Appendix~\ref{app:train_and_inf}.
According to Table~\ref{tab:ablate_order}, we can safely conclude that: benefiting from the well-designed structure, different orders does not significantly influence the eventual performance~(even with Random method which evidently enlarges the search space).
Furthermore, we can take advantage of the self-adaptive node selection and mark AST nodes with decoding timestamps during inference. It provides better symbolic interpretation on the decoding process. Through case studies~(partly illustrated in Table~\ref{tab:case_order}), we find that in roughly $76.2\%$ cases, the model prefers to expand node ``{\tt from}'' before ``{\tt select}'' where only one table is required in the target SQL. This observation is consistent with previous literature~\cite{bridge} that prioritizing \textsc{From} clause according to the execution order is superior to the written order that \textsc{Select} clause comes first. However, this ratio decreases to $13.9\%$ when it encounters the JOIN of multiple tables. In this situation, we speculate that the model tends to identify the user intents~(\textsc{Select} clause) and requirements~(\textsc{Where} clause) first according to the input question, and switch to the complicated construction of a connected table view based on the database schema graph. More examples are provided in Appendix~\ref{app:case_order}.
\begin{table}[htbp]
  \centering
  \resizebox{0.48\textwidth}{!}{
    \begin{tabular}{l}
    \hline

    \hline

    \textbf{DB:} {\it concert\_singer} \\
    \textbf{Question:} How many singers do we have? \\
    \textbf{SQL:} {\tt select} {\tt count(*)} \colorbox{sqlcolor}{{\tt from} \emph{singer}} \\
    \textbf{SQL AST:} \\
{\tt sql} $\rightarrow$ Node[ $j=0$, {\tt sql} := SQL({\tt from}, {\tt select}, {\tt condition}, {\tt groupby}, {\tt orderby}) ]\\
\qquad {\tt groupby} $\rightarrow$ Node[ $j=1$, {\tt groupby} := NoGroupBy() ]\\
\qquad \colorbox{fromcolor}{{\tt from} $\rightarrow$ Node[ $j=2$, {\tt from} := FromTableOne({\tt tab\_id}, {\tt condition}) ]}\\
\qquad\qquad {\tt tab\_id} $\rightarrow$ Leaf[ $j=3$, {\tt tab\_id} := \emph{singer} ]\\
\qquad\qquad {\tt condition} $\rightarrow$ Node[ $j=4$, {\tt condition} := NoCondition() ]\\
\qquad {\tt condition} $\rightarrow$ Node[ $j=5$, {\tt condition} := NoCondition() ]\\
\qquad \colorbox{selectcolor}{{\tt select} $\rightarrow$ Node[ $j=6$, {\tt select} := SelectColumnOne({\tt distinct}, {\tt col\_unit}) ]}\\
\qquad\qquad {\tt distinct} $\rightarrow$ Node[ $j=7$, {\tt distinct} := False() ]\\
\qquad\qquad {\tt col\_unit} $\rightarrow$ Node[ $j=8$, {\tt col\_unit} := UnaryColumnUnit({\tt agg\_op}, {\tt distinct}, {\tt col\_id}) ]\\
\qquad\qquad\qquad {\tt distinct} $\rightarrow$ Node[ $j=9$, {\tt distinct} := False() ]\\
\qquad\qquad\qquad {\tt agg\_op} $\rightarrow$ Node[ $j=10$, {\tt agg\_op} := Count() ]\\
\qquad\qquad\qquad {\tt col\_id} $\rightarrow$ Leaf[ $j=11$, {\tt col\_id} := * ]\\
\qquad {\tt orderby} $\rightarrow$ Node[ $j=12$, {\tt orderby} := NoOrderBy() ]\\
    \hdashline
    \textbf{DB:} {\it poker\_player} \\
    \textbf{Question:} What are the names of poker players? \\
    \textbf{SQL:} {\tt select} \emph{T1.Name} \colorbox{sqlcolor}{{\tt from} \emph{people} {\tt as} \emph{T1} {\tt join} \emph{poker\_player} {\tt as} \emph{T2} {\tt on} \emph{T1.People\_ID} = \emph{T2.People\_ID}} \\
    \textbf{SQL AST:} \\
{\tt sql} $\rightarrow$ Node[ $j=0$, {\tt sql} := SQL({\tt from}, {\tt select}, {\tt condition}, {\tt groupby}, {\tt orderby}) ]\\
\qquad \colorbox{selectcolor}{{\tt select} $\rightarrow$ Node[ $j=1$, {\tt select} := SelectColumnOne({\tt distinct}, {\tt col\_unit}) ]}\\
\qquad\qquad {\tt distinct} $\rightarrow$ Node[ $j=2$, {\tt distinct} := False() ]\\
\qquad\qquad {\tt col\_unit} $\rightarrow$ Node[ $j=3$, {\tt col\_unit} := UnaryColumnUnit({\tt agg\_op}, {\tt distinct}, {\tt col\_id}) ]\\
\qquad\qquad\qquad {\tt agg\_op} $\rightarrow$ Node[ $j=4$, {\tt agg\_op} := None() ]\\
\qquad\qquad\qquad {\tt distinct} $\rightarrow$ Node[ $j=5$, {\tt distinct} := False() ]\\
\qquad\qquad\qquad {\tt col\_id} $\rightarrow$ Leaf[ $j=6$, {\tt col\_id} := \emph{people.Name} ]\\
\qquad {\tt condition} $\rightarrow$ Node[ $j=7$, {\tt condition} := NoCondition() ]\\
\qquad {\tt groupby} $\rightarrow$ Node[ $j=8$, {\tt groupby} := NoGroupBy() ]\\
\qquad \colorbox{fromcolor}{{\tt from} $\rightarrow$ Node[ $j=9$, {\tt from} := FromTableTwo({\tt tab\_id}, {\tt tab\_id}, {\tt condition}) ]}\\
\qquad\qquad {\tt tab\_id} $\rightarrow$ Leaf[ $j=10$, {\tt tab\_id} := \emph{poker\_player} ]\\
\qquad\qquad {\tt tab\_id} $\rightarrow$ Leaf[ $j=11$, {\tt tab\_id} := \emph{people} ]\\
\qquad\qquad {\tt condition} $\rightarrow$ Node[ $j=12$, {\tt condition} := CmpCondition({\tt col\_unit}, {\tt cmp\_op}, {\tt value}) ]\\
\qquad\qquad\qquad {\tt col\_unit} $\rightarrow$ Node[ $j=13$, {\tt col\_unit} := UnaryColumnUnit({\tt agg\_op}, {\tt distinct}, {\tt col\_id}) ]\\
\qquad\qquad\qquad\qquad {\tt distinct} $\rightarrow$ Node[ $j=14$, {\tt distinct} := False() ]\\
\qquad\qquad\qquad\qquad {\tt agg\_op} $\rightarrow$ Node[ $j=15$, {\tt agg\_op} := None() ]\\
\qquad\qquad\qquad\qquad {\tt col\_id} $\rightarrow$ Leaf[ $j=16$, {\tt col\_id} := \emph{poker\_player.People\_ID} ]\\
\qquad\qquad\qquad {\tt cmp\_op} $\rightarrow$ Node[ $j=17$, {\tt cmp\_op} := Equal() ]\\
\qquad\qquad\qquad {\tt value} $\rightarrow$ Node[ $j=18$, {\tt value} := ColumnValue({\tt col\_id}) ]\\
\qquad\qquad\qquad\qquad {\tt col\_id} $\rightarrow$ Leaf[ $j=19$, {\tt col\_id} := \emph{people.People\_ID} ]\\
\qquad {\tt orderby} $\rightarrow$ Node[ $j=20$, {\tt orderby} := NoOrderBy() ]\\
    \hline

    \hline
    \end{tabular}%
    }
   \caption{Case study on traversal order using method DFS+Random on Spider~\cite{spider}. $j$ represents the decoding timestep when the node is expanded.}
  \label{tab:case_order}%
\end{table}%

\section{Related Work}
\label{sec:related}
\paragraph{Text-to-SQL Decoding}
SQL programs exhibit strict syntactic and semantic constraints. Without restriction, the decoder will waste computation in producing ill-formed programs. To tackle this \emph{constrained decoding} problem,
existing methods can be classified into two categories: \emph{token-based} and \emph{grammar-based}.
1) A token-based decoder~\cite{bridge,picard} directly generates each token in the SQL autoregressively.
This method 
requires syntax- or execution-guided decoding~\cite{execution-guided} to periodically eliminate invalid partial programs from the beam during inference. 2) A grammar-based decoder~\cite{type-constraints,syntactic,irnet} predicts a sequence of actions~(or grammar rules) to construct the equivalent SQL AST instead. Type constraints~\cite{type-constraints} can be easily incorporated into node types and grammar rules. In this branch, both top-down (IRNet,~\citealp{irnet}) and bottom-up (\textsc{SmBoP},~\citealp{smbop}) grammars have been applied to instruct the decoding. We focus on the grammar-based category and propose an AST decoder which is compatible with Transformer.

\paragraph{Structure-aware Decoding}
Various models have been proposed for structured tree generation, such as \textsc{Asn}~\cite{asn}, \textsc{TranX}~\cite{tranx}, and \textsc{TreeGen}~\cite{treegen}. These methods introduce complex shortcut connections or specialized modules to model the structure of output trees. In contrast, StructCoder~\cite{structcoder} still adopts the token-based paradigm and devises two auxiliary tasks~(AST paths and data flow prediction) during training to encode the latent structure into objective functions. \citet{peng2021integrating} also integrates tree path encodings into attention module, but they work on the encoding part and assume that the complete AST is available. In this work, we propose a succinct ASTormer architecture for the decoder and construct the SQL AST on-the-fly during inference.

\section{Conclusion}
\label{sec:conclusion}
In this work, we propose an AST structure-aware decoder for text-to-SQL. It integrates the absolute and relative position of each node in the AST into the Transformer decoder.
Extensive experiments verify that ASTormer is more effective and efficient than traditional LSTM-series AST decoder. And it can be easily adapted to more flexible traversing orders.
Future work includes extending ASTormer to more general situations like graph generation.
\section*{Limitations}
In this work, we mainly focus on grammar-based decoders for text-to-SQL. Both token-based parsers equipped with relatively larger pre-trained language models such as T5-3B~\cite{t5} and prompt-based in-context learning methods with large language models such as ChatGPT~\cite{instructgpt} are beyond the scope, since this paper is restricted to local interpretable small-sized models~(less than 1B parameters) which are cheaper and faster.



\bibliography{custom}
\bibliographystyle{acl_natbib}

\appendix

\clearpage
\section{Grammar Rules}
\label{app:grammar}
In Table~\ref{tab:grammar}, we provide the complete set of grammar rules for SQL AST parsing. This specification conforms to a broader category called \emph{abstract syntax description language}~(ASDL,~\citealp{asdl}). Take the following grammar rule as an example:
\begin{multline*}
    {\tt select} := \\
    \text{SelectOneColumn}({\tt distinct}, {\tt col\_unit}),
\end{multline*}
the corresponding \textsc{ApplyRule} action will expand the non-terminal node of type ``{\tt select}''. Concretely, two unexpanded children nodes of types ``{\tt distinct}'' and ``{\tt col\_unit}'' will be attached to the parent node of type ``{\tt select}''.

\section{Construction of Relations On-the-fly}
\label{app:on-the-fly}
In this part, we introduce how to construct the relation set $Z_j=\{z_{ji}\}_{i=1}^j$ at timestep $j$ given all previous relation sets $Z_{<j}$ during inference. The entire set $Z_j$ can be divided into three parts:
\begin{align*}
Z_j=&Z_j^0\bigcup Z_j^1\bigcup Z_j^2,\\
=&\{z_{jj}\}\bigcup \{z_{ji}\}_{i=1}^{p_j}\bigcup \{z_{ji}\}_{i=p_j+1}^{j-1},
\end{align*}
where $p_j$ denotes the timestep when the parent of the current frontier node $n_j$ is expanded~($p_j<j$).
\begin{itemize}
    \item For $Z_j^0=\{z_{jj}\}$, we directly use $z_{jj}=(0, 0)$.
    \item For each $z_{ji}\in Z_j^1=\{z_{ji}\}_{i=1}^{p_j}$, we retrieve the corresponding relation $z_{p_ji}\in Z_{p_j}$ with the same index $i$. Assume that $z_{p_ji}=(k,s)$, then $z_{ji}$ must be
    $$z_{ji}=(\text{clamp}(k+1, R), s).$$
    \item For each $z_{ji}\in Z_j^2=\{z_{ji}\}_{i=p_j+1}^{j-1}$, although $z_{p_ji}\notin Z_{p_j}$, the counter-part $z_{ip_j}$ must be already constructed in $Z_i$ with $p_j<i<j$. Assume that $z_{ip_j}=(k,s)$, considering the symmetric definition of relations, $z_{ji}$ must be
    $$z_{ji}=(\text{clamp}(s+1, R), k).$$
\end{itemize}
Based on the three rules above, the relation set $Z_j$ at each timestep can be efficiently computed with little overheads. An empirical comparison with LSTM-based AST decoder on inference time is presented in Table~\ref{tab:inference}.
\begin{table}[htbp]
  \centering
    \begin{tabular}{c|ccc}
    \hline

    \hline
    \multicolumn{1}{c|}{\textbf{Dataset}} & \textbf{Spider} & \textbf{SParC} & \textbf{CoSQL} \\
    \hline
    LSTM & 206.6  & 191.5  & 201.0  \\
    ASTormer & 237.0  & 200.7  & 199.1  \\
    \hline

    \hline
    \end{tabular}%
  \caption{Inference time comparison~(seconds/per $1000$ samples under the same configuration).}
  \label{tab:inference}%
\end{table}%

\begin{table*}[htbp]
  \centering
  \resizebox{0.98\textwidth}{!}{
    \begin{tabular}{ll}
    \hline
    \hline
     & \\
     &\textbf{\large Terminal types:} \\
     & \\
     &\quad {\tt tab\_id}, {\tt col\_id}, {\tt tok\_id}\\
     & \\
     &\textbf{\large Grammar rules for non-terminal types:} \\
     & \\
     &\quad {\tt sql} := Intersect({\tt sql}, {\tt sql}) $|$ Union({\tt sql}, {\tt sql}) $|$ Except({\tt sql}, {\tt sql}) $|$ SQL({\tt from}, {\tt select}, {\tt condition}, {\tt groupby}, {\tt orderby})\\
     & \\
     &\quad {\tt select} := SelectOneColumn({\tt distinct}, {\tt col\_unit}) $|$ SelectTwoColumn({\tt distinct}, {\tt col\_unit}, {\tt col\_unit}) $|$ $\cdots$\\
     & \\
     &\quad {\tt from} := FromOneSQL({\tt sql}) $|$ FromTwoSQL({\tt sql}, {\tt sql}) $|$ $\cdots$ \\
     &\quad\qquad\qquad $|$ FromOneTable({\tt tab\_id}) $|$ FromTwoTable({\tt tab\_id}, {\tt tab\_id}, {\tt condition}) $|$ $\cdots$ \\
     & \\
     &\quad {\tt groupby} := NoGroupBy \\
     &\quad\qquad\qquad $|$ GroupByOneColumn({\tt col\_id}, {\tt condition}) $|$ GroupByTwoColumn({\tt col\_id}, {\tt col\_id}, {\tt condition}) $|$ $\cdots$ \\
     & \\
     &\quad {\tt orderby} := NoOrderBy \\
     &\quad\qquad\qquad $|$ OrderByOneColumn({\tt col\_unit}, {\tt order}) $|$ OrderByTwoColumn({\tt col\_unit}, {\tt col\_unit}, {\tt order}) $|$ $\cdots$ \\
     &\quad\qquad\qquad $|$ OrderByLimitOneColumn({\tt col\_unit}, {\tt order}, {\tt tok\_id}) $|$ OrderByLimitTwoColumn({\tt col\_unit}, {\tt col\_unit}, {\tt order}, {\tt tok\_id}) $|$ $\cdots$ \\
     & \\
     &\quad {\tt order} := Asc $|$ Desc \\
     & \\
     &\quad {\tt condition} := NoCondition \\
     &\quad\qquad\qquad $|$ AndTwoCondition({\tt condition}, {\tt condition}) $|$ AndThreeCondition({\tt condition}, {\tt condition}, {\tt condition}) $|$ $\cdots$ \\
     &\quad\qquad\qquad $|$ OrTwoCondition({\tt condition}, {\tt condition}) $|$ OrThreeCondition({\tt condition}, {\tt condition}, {\tt condition}) $|$ $\cdots$ \\ 
     &\quad\qquad\qquad $|$ BetweenCondition({\tt col\_unit}, {\tt value}, {\tt value})\\
     &\quad\qquad\qquad $|$ CmpCondition({\tt col\_unit}, {\tt cmp\_op}{\tt value})\\
     & \\
     &\quad {\tt cmp\_op} := Equal $|$ NotEqual $|$ GreaterThan $|$ GreaterEqual $|$ LessThan $|$ LessEqual $|$ Like $|$ NotLike $|$ In $|$ NotIn \\
     & \\
     &\quad {\tt value} := SQLValue({\tt sql}) $|$ LiteralValue({\tt tok\_id}) $|$ ColumnValue({\tt col\_id}) \\
     & \\
     &\quad {\tt col\_unit} := UnaryColumnUnit({\tt agg\_op}, {\tt distinct}, {\tt col\_id}) $|$ BinaryColumnUnit({\tt agg\_op}, {\tt unit\_op}, {\tt col\_id}, {\tt col\_id}) \\
     & \\
     &\quad {\tt distinct} := True $|$ False \\
     & \\
     &\quad {\tt agg\_op} := None $|$ Max $|$ Min $|$ Count $|$ Sum $|$ Avg \\
     & \\
     &\quad {\tt unit\_op} := Minus $|$ Plus $|$ Times $|$ Divide \\
     & \\
    \hline
    \hline
    \end{tabular}%
    }
  \caption{The checklist of all grammar rules used throughout all experiments. ``$|$'' is the separator for different rule constructors. ``$\cdots$'' means part of the quantitive grammar rules are omitted which can be easily inferred, and the maximum number of children types is determined by statistics of the training set. Note that grammar rules such as ``{\tt order} := Asc $|$ Desc'' denote no children node will be attached to the parent.}
  \label{tab:grammar}%
\end{table*}%

\section{Details of Different Traversal Paths}
\label{app:alg}
\subsection{Modeling of AST Construction}
To construct the SQL AST $y^a$ in an auto-regressive fashion, there are multiple top-down traversal paths $\mathcal{T}(y^a)$. Each traversal path $\rho \in\mathcal{T}(y^a)$ can be further decoupled into the sequence of (AST node, output action) pairs $\rho =\bigl((n_1, a_1),\cdots,(n_{|y^a|}, a_{|y^a|})\bigl)$. The probabilistic modeling is formulated as:
\begin{align*}
\log P(y^a)=&\sum_{\rho\in\mathcal{T}(y^a)}\log P(\rho)\\
=&\sum_{\rho\in\mathcal{T}(y^a)}\log P(n_1, a_1,\cdots,n_{|y^a|}, a_{|y^a|})\\
=&\sum_{\rho\in\mathcal{T}(y^a)}\sum_{j=1}^{|y^a|}\log P(n_j, a_j|n_{<j},a_{<j})\\
=&\sum_{\rho\in\mathcal{T}(y^a)}\sum_{j=1}^{|y^a|}\bigl(\log P(n_j|n_{<j},a_{<j})\\
&\qquad\qquad\quad + \log P(a_j|n_{\le j},a_{<j})\bigl),
\end{align*}
where the conditioning on input $\mathbf{X}$ is omitted for brevity.
Given a specific traversal path $\rho$, the modeling of $\log P(n_j,a_j|n_{<j},a_{<j})$ at each timestep $j$ is divided into two parts:
$$\underbrace{\log P(n_j|n_{<j},a_{<j})}_{\text{selection of frontier node }n_j} + \underbrace{\log P(a_j|n_{\le j}, a_{<j})}_{\text{modeling of output action }a_j}.$$
The first part gives the distribution of choosing one unexpanded node in $y^a_{j-1}$, while the second part is exactly the output of the AST decoder.

Most previous work simplifies the first part by assuming one canonical traversal path~(DFS+L2R) given AST $y^a$. In other words, $\mathcal{T}_{\text{dfs}}^{\text{l2r}}(y^a)$ is a singleton set containing merely one unrolled path, and $P(n_j|n_{<j},a_{<j})$ is a unit mass distribution where only the leftmost node in the frontier node set restricted by DFS has the probability $1$~(similarly for $\mathcal{T}_{\text{bfs}}^{\text{l2r}}(y^a)$). In this case, we only need to care about the distribution of output actions since each frontier node $n_j$ is uniquely determined by $y^a_{j-1}$. It gives us exactly the training objective in Eq.(\ref{eq:loss}):
\begin{align*}
\log P(y^a)=&\sum_{j=1}^{|y^a|}\log P(a_j|n_{<j}, a_{<j})\\
=&\sum_{j=1}^{|y^a|}\log P(a_j|a_{<j}).
\end{align*}

\begin{figure}[htbp]
    \centering
    \includegraphics[width=0.49\textwidth]{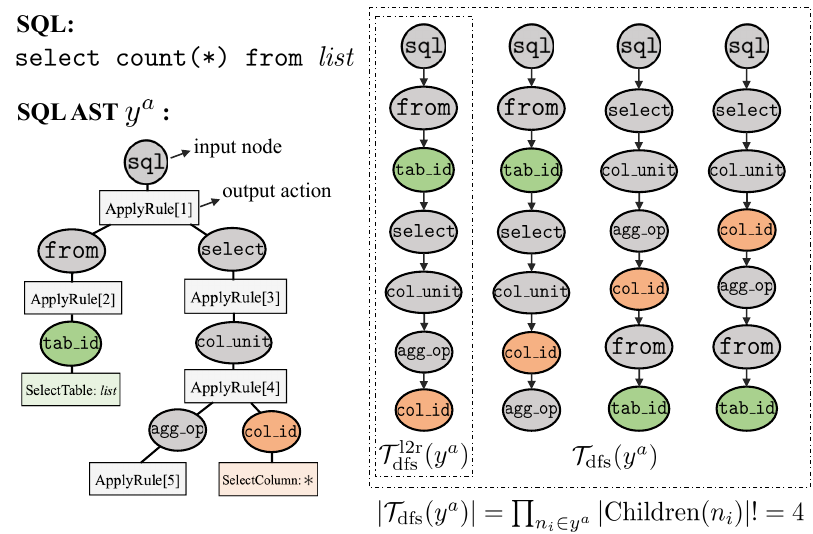}
    \caption{Illustration of different traversal paths $\mathcal{T}_{\text{dfs}}(y^a)$ re-using the same example in Figure~\ref{fig:astormer}.}
    \label{fig:paths}
\end{figure}

In this work, we relax the permutation prior injected in the ``horizontal'' direction~(not necessarily L2R) and propose the Random method~(\cref{sec:order}). Concretely, the entire set of acceptable traversal paths is restricted by DFS~(or BFS), and the total number of paths in $\mathcal{T}_{\text{dfs}}(y^a)$~(or $\mathcal{T}_{\text{bfs}}(y^a)$) equals to $$|\mathcal{T}_{\text{dfs}}(y^a)|=|\mathcal{T}_{\text{bfs}}(y^a)|=\prod_{n_i\in y^a}|\text{Children}(n_i)|!\ ,$$
where $|\text{Children}(n_i)|$ denotes the number of children for each node $n_i$ in AST $y^a$. An illustration of different traversal paths $\mathcal{T}_{\text{dfs}}(y^a)$ and its cardinality is provided in Figure~\ref{fig:paths}. Assume that each path $\rho$ has equal prior, then the selection of frontier node $P(n_j|n_{<j},a_{<j})$ is a uniform distribution over the current frontier node set restricted by DFS~(or BFS). And the computation of $\log P(y^a)$ can be reduced to: (take DFS as an example)
\begin{align*}
\log P(y^a)=&\sum_{\rho\in\mathcal{T}_{\text{dfs}}(y^a)}\Bigl(\sum_{j=1}^{|y^a|}\log P(n_j|n_{<j}, a_{<j})\\
&\qquad\quad+\sum_{j=1}^{|y^a|}\log P(a_j|n_{\le j}, a_{<j})\Bigl)\\
=&\sum_{\rho\in\mathcal{T}_{\text{dfs}}(y^a)}\Bigl(\log\frac{1}{\prod_{n_i\in y^a}|\text{Children}(n_i)|!}\\
&\qquad\quad+\sum_{j=1}^{|y^a|}\log P(a_j|n_{\le j}, a_{<j})\Bigl)\\
=&c+\sum_{\rho\in\mathcal{T}_{\text{dfs}}(y^a)}\sum_{j=1}^{|y^a|}\log P(a_j|n_{\le j}, a_{<j}),
\end{align*}
where $c$ is a constant value irrelevant of $\rho$ and only dependent on the shape of $y^a$. Considering enormous traversal paths in $\mathcal{T}_{\text{dfs}}(y^a)$, we adopt uniform sampling during training and enlarge the search space over frontier nodes $n_j$ during inference~(detailed in Appendix~\ref{app:train_and_inf}).

\subsection{Training and Inference Algorithms}
\label{app:train_and_inf}
\begin{algorithm*}[htbp]
\caption{Generic Training Algorithm}
\label{alg:pita}
\begin{algorithmic}[1]
\Require Labeled AST $y^a$; Encoder output $\mathbf{X}=[\mathbf{Q};\mathbf{T};\mathbf{C}]$; Order method $m$;
\Ensure Training loss $\mathcal{L}$ for one sample;

\State{traverse AST $y^a$ via DFS/BFS+L2R and record necessary information for each node, including traverse timestep $j$, node type $n_j^{\tau}$, parent grammar rule $r_{p_j}$, node depth $D^a(n_j)$ and output action $a_j$;}
\State{construct the relational matrix $Z=\{z_{ji}\}_{1\le i, j\le |y^a|}$ for each node pair in $y^a$;}
\\
\If{$m\in\{\text{DFS+Random, BFS+Random}\}$}
{\color{darkblue}\Comment{only ``Random'' needs sampling and permutation}}
\State{sample a traversal path over $y^a$ following $m$ and record timestamps;}
\State{permute the $1$-dimension sequence of $(n_j^{\tau}, r_{p_j}, D^a(n_j), a_j)$ based on timestamps;}
\State{permute the relation matrix $Z$ twice in two dimensions based on timestamps;}
\EndIf
\\
\State{compute training loss $\mathcal{L}$ in parallel given $\mathbf{X}$, the current traversal sequence and relation matrix $Z$;}
\State{\Return training loss $\mathcal{L}$}
\end{algorithmic}
\end{algorithm*}

\begin{algorithm*}[htbp]
\caption{Generic Inference Algorithm}
\label{alg:piia}
\begin{algorithmic}[1]
\Require Encoder output $\mathbf{X}=[\mathbf{Q};\mathbf{T};\mathbf{C}]$; Beam size $K$; Order method $m$; Max decoding steps $M$;
\Ensure Predicted SQL program $\hat{y}$;

\State{initialize AST $y_0^a$ with a single node of root type ``{\tt sql}'';}
\State{initialize beam $\mathcal{B}_0=\{y^a_0\}$ with one hypothesis;}
\State{initialize the set of completed hypotheses $\mathcal{H}=\emptyset$}

\For{$j=1$ to $M$}
\State{$\mathcal{N}\ =\ \emptyset$ ;}
\For{each $y^a_{j-1}$ in $\mathcal{B}_{j-1}$}\Comment{in the \emph{symbolic space}}
\If{$m\in\{\text{DFS+L2R, BFS+L2R}\}$}{\color{darkblue}\Comment{L2R: directly follow the canonical order}}
\State{find the leftmost node $n_j$ in the frontier node set of $y^a_{j-1}$ ;}
\State{$\mathcal{N}\ =\ \mathcal{N}\cup\{(y^a_{j-1}, n_j)\}$ ;}
\Else{\color{darkblue}\Comment{Random: enlarge the search space with all feasible options}}
\For{each distinct unexpanded node $n_{j}$ in the frontier node set of $y^a_{j-1}$}
\State{$\mathcal{N}\ =\ \mathcal{N}\cup\{(y^a_{j-1}, n_j)\}$ ;}
\EndFor
\EndIf
\EndFor

\State{$\mathcal{A}\ =\ \emptyset$ ;}
\For{each $(y_{j-1}^a, n_j)$ in $\mathcal{N}$}\Comment{in the \emph{neural space}}
\State{$\boldsymbol{n}_j\ =\ $\textsc{ProcessInputs}$(y^a_{j-1}$, $n_j$, $\mathbf{X})$~(see \cref{sec:ape} and \cref{sec:rpe}) ;}
\State{compute $P(a_j|\boldsymbol{n}_j)$ based on node type $n_j^{\tau}$~(see \cref{sec:prob}) ;}

\For{each $a_{j}$ that satisfies node type constraints}
\State{$\mathcal{A}\ =\ \mathcal{A}\cup\{(y^a_{j-1}, n_j, a_j)\}$ ;}
\EndFor
\EndFor
\State{sort and retain top-$K$ $(y_{j-1}^a, n_j, a_j)$ triples in $\mathcal{A}$;}
{\Comment{score of $y^a_{j}$ is sum of logprobs $P(a_{\le j})$}}
\State{$\mathcal{B}_{j}\ =\ \emptyset$ ;}
\For{each $(y^a_{j-1}, n_j, a_j)$ in $\mathcal{A}$}\Comment{in the \emph{symbolic space}}
\State{$y^a_{j}\leftarrow$\textsc{ApplyAction}$(y^a_{j-1},n_j,a_j)$ ;}
\State{update the frontier node set for $y^a_{j}$ via order method $m$ ;}
{\color{darkblue}{\Comment{DFS or BFS according to $m$}}}
\If{$y^a_j$ is completed}
\State{$\mathcal{H}\ =\ \mathcal{H}\cup \{y^a_j\}$ ;}
\Else
\State{$\mathcal{B}_j\ =\ \mathcal{B}_j\cup \{y^a_j\}$ ;}
\EndIf
\EndFor
\EndFor
\State{$\hat{y}^a\ =\ $AST with the maximum score in $\mathcal{H}$ ;}
\State{convert SQL AST $\hat{y}^a$ to SQL program $\hat{y}$ ;}
\State{\Return $\hat{y}$}
\end{algorithmic}
\end{algorithm*}
The generic training and inference algorithms are provided in Alg.\ref{alg:pita} and Alg.\ref{alg:piia}, respectively. We extend the traditional DFS+L2R decoding order into more flexible choices according to the order method $m$~(DFS/BFS + L2R/Random).

For training, we firstly record all necessary information in one pass following the canonical traversal order DFS/BFS+L2R~(line $1$-$2$ in Alg.\ref{alg:pita}). During this traversal, each node is also labeled with the timestamp when it is visited. Next, for each training iteration, if the order method L2R is utilized, we re-use the same traversal path to calculate the training loss for each data points~(line $10$). Otherwise, if the order method Random is required, we sample one traversal path over $y^a$ following $m$ and extract the timestamp for each frontier node~(line $5$ in Alg.\ref{alg:pita}). Treating these timestamps as indexes, we can sort the input-output sequence of (frontier node, output action) pairs as well as the $2$-dimension relational matrix $Z$~(sort twice), line $6$-$7$ in Alg.~\ref{alg:pita}.

As for inference, the decoding method $m$ should be consistent with that during training. Concretely, if the canonical method L2R is adopted during training, we directly choose the leftmost node in the frontier node set at each timestep~(line $8$-$9$ in Alg.~\ref{alg:piia}). Otherwise for order method Random, to avoid stochasticity and enlarge the search space, we enumerate all distinct nodes in the frontier node set into the decoder as inputs~(line $11$-$13$ in Alg.\ref{alg:piia}). Besides, when applying action $a_j$ to expand the frontier node $n_j$, we also utilize DFS or BFS to restrict the frontier node set depending on the order method $m$~(line $28$ in Alg.\ref{alg:piia}).

\section{Datasets}
\label{app:dataset}
Experiments are conducted on five cross-domain multi-table text-to-SQL benchmarks, namely Spider~\cite{spider}, SParC~\cite{sparc}, CoSQL~\cite{cosql}, DuSQL~\cite{dusql} and Chase~\cite{chase}. The first three are English benchmarks, while the latter two are in Chinese. In these benchmarks, the databases of training, validation, and test sets do not overlap. Note that, SParC, CoSQL and Chase are context-dependent. Each training sample is an user-system interaction which contains multiple turns. For contextual scenarios, we directly append history utterances to the current one separated by a delimiter ``|" and treat all utterances as the entire question $Q$. The official evaluation metrics for each dataset are reported on the development set. We exclude the {\tt baseball\_1} database from the training data because the schema of this database is too large. Statistics are listed in Table \ref{tab:dataset}.
\begin{table}[htbp]
  \centering
    \resizebox{0.49\textwidth}{!}{
    \begin{tabular}{l|ccccc}
    \hline
    
    \hline
    \multicolumn{1}{c|}{\textbf{Dataset}} & \textbf{Spider} & \textbf{SParC} & \textbf{CoSQL} & \textbf{DuSQL} & \textbf{Chase} \\
    \hline 
    Language & EN & EN & EN & ZH & ZH \\
    \# samples & 10,181  & 12,726 & 15,598 & 28,762 & 17,940 \\
    Avg \# tables per DB & 5.1 & 5.3 & 5.4 & 4.0 & 4.6 \\
    Multi-turn & \XSolidBrush  & {\large\checkmark}  & {\large\checkmark} & \XSolidBrush & {\large\checkmark} \\
    \hline 
    Avg \# $Q$ tokens & 13.9 & 20.6 & 32.6 & 31.9 & 31.4 \\
    Avg \# tables & 6.2 & 5.5 & 5.7 & 4.3 & 4.9 \\
    Avg \# columns & 29.6 & 28.5 & 29.3 & 25.3 & 25.9 \\
    \hline 
    Avg \# SQL tokens & 37.3 & 28.8 & 30.4 & 61.7 & 45.9 \\
    Avg \# AST nodes & 32.4 & 25.7 & 27.0 & 34.1 & 30.0 \\
    \hline 
    Metrics & EM, EX & QM, IM & QM, IM & EM & QM, IM \\
    \hline

    \hline
    \end{tabular}%
}
  \caption{Statistics of all benchmarks. DB: database.}
    \label{tab:dataset}%
\end{table} %

\subsection{Evaluation Metrics for Text-to-SQL}
\paragraph{Exact Set Match~(EM)} This metric measures the equivalence of two SQL queries by comparing each component. The prediction is correct only if each fine-grained clause or unit is correct. Order issues will be ignored, such that ``{\tt SELECT} {\tt col1,} {\tt col2}" equals ``{\tt SELECT} {\tt col2,} {\tt col1}". However, EM only checks the SQL sketch and ignores SQL values.
\paragraph{Execution Accuracy~(EX)} It measures the accuracy by comparing the execution results instead. However, erroneous SQLs~(called \emph{spurious programs}) may happen to attain the same correct results on a particular database. To alleviate this problem, Zhong et al.~\cite{testsuite} proposed a distilled test suite that contains multiple databases for each domain. The predicted SQL is correct only if the execution results are equivalent to the ground truth on all these databases.
\paragraph{Question-level Exact Set Match~(QM)} It is exactly the same with the EM metric if we treat each turn in one interaction as a single (question, SQL) sample. In other words, the interaction-level sample is ``flattened" into multiple question-level samples.
\paragraph{Interaction-level Exact Set Match~(IM)} It reports the accuracy at the interaction level. One interaction is correct if the results of all turns are right on the metric EM.

\section{Results on Chinese Benchmarks}
\label{app:chinese}
To verify the universality of ASTormer under different languages, we also experiment on two multi-table cross-domain Chinese benchmarks, namely single-turn DuSQL~\cite{dusql} and multi-turn Chase~\cite{chase}. Due to the scarcity of baselines on these two datasets, we also implement another competitive system, a token-based text-to-SQL parser which integrates the graph encoder RATSQL and the schema copy mechanism. Performances are presented in Table~\ref{tab:dusql} and~\ref{tab:chase}.
\begin{table}[htbp]
  \centering
  \resizebox{0.48\textwidth}{!}{
    \begin{tabular}{c|c|cc}
    \hline

    \hline
    \textbf{Category} & \textbf{Method} & \textbf{EM} & \textbf{EM w/ val} \\
    \hline
    token-based & Seq2Seq+Copy~\cite{see-etal-2017-get} & 6.6   & 2.6  \\
    \hdashline
    \multirow{2}{*}{\textbf{Ours}} & RATSQL + w/ LSTM & 78.7  & 61.4  \\
          & RATSQL + w/ Transformer & 79.0  & 62.5  \\
    \hline
    \multirow{3}{*}{grammar-based} & SyntaxSQLNet~\cite{syntaxsqlnet} & 14.6  & 7.1  \\
          & IRNet~\cite{irnet} & 38.4  & 18.4  \\
          & IRNetExt~\cite{dusql} & 59.8  & 56.2  \\
    \hdashline
    \multirow{2}{*}{\textbf{Ours}} & RATSQL + w/ LSTM & 79.3  & 62.9  \\
          & RATSQL + w/ ASTormer & \textbf{79.7} & \textbf{64.4} \\
    \hline

    \hline
    \end{tabular}%
    }
  \caption{Main results on the validation set of Chinese benchmark DuSQL~\cite{dusql}. For fair comparison, all methods above do not utilize PLMs.}
  \label{tab:dusql}%
\end{table}%

\begin{table}[htbp]
  \centering
  \resizebox{0.48\textwidth}{!}{
    \begin{tabular}{c|c|cc}
    \hline

    \hline
    \textbf{Category} & \textbf{Method} & \textbf{QM} & \textbf{IM} \\
    \hline
    \multirow{2}{*}{token-based} & EditSQL~\cite{editsql} & 37.7  & 17.4  \\
          & IGSQL~\cite{igsql} & 41.4  & 20.0  \\
    \hdashline
    \multirow{2}{*}{\textbf{Ours}} & RATSQL + w/ LSTM & 45.7  & 20.5  \\
          & RATSQL + w/ Transformer & 45.6  & 19.1  \\
    \hline
    grammar-based & DuoRAT~\cite{duorat} & 35.1  & 14.6  \\
    \hdashline
    \multirow{2}{*}{\textbf{Ours}} & RATSQL + w/ LSTM & 46.7  & 21.5  \\
          & RATSQL + w/ ASTormer & \textbf{47.1} & \textbf{22.4} \\
    \hline

    \hline
    \end{tabular}%
    }
  \caption{Main results on the validation set of Chinese multi-turn benchmark Chase~\cite{chase}. The default PLM is BERT~\cite{chinese-bert-wwm}.}
  \label{tab:chase}%
\end{table}%

Accordingly, we observe that the two basic discoveries are in accordance with those on English benchmarks. 1) The proposed ASTormer architecture steadily outperforms traditional LSTM-series networks. And we establish new state-of-the-art performances on two Chinese multi-table cross-domain datasets using the same PLMs~(or without). 2) For models on the same scale, grammar-based AST decoders are superior to pure token-based decoders without generative pre-training, which is largely attributed to the explicit strucural modeling.

\section{Case Study on Traversal Order}
\label{app:case_order}
In this part, we provide more examples on the validation set of Spider~\cite{spider} marked up with decoding timesteps during inference with order DFS+Random~(Table~\ref{tab:case_order1} to \ref{tab:case_order2}). In a nutshell, the expanding order of AST nodes is affected by the interplay of multiple factors, including the surface form of the user question, the structure of input database schema, syntactic constraints of grammar rules, and the intrinsic stochasticity triggered by the training process.
\begin{table*}[htbp]
  \centering
  \resizebox{0.9\textwidth}{!}{
    \begin{tabular}{l}
    \hline

    \hline
    \textbf{DB:} {\it concert\_singer} \\
    \textbf{Question:} What is the total number of singers? \\
    \textbf{SQL:} {\tt select} {\tt count} ( \emph{*} ) {\tt from} \emph{singer} \\
    \textbf{SQL AST:} \\
{\tt sql} $\rightarrow$ Node[ $j=0$, {\tt sql} := SQL({\tt from}, {\tt select}, {\tt condition}, {\tt groupby}, {\tt orderby}) ]\\
\qquad {\tt groupby} $\rightarrow$ Node[ $j=1$, {\tt groupby} := NoGroupBy() ]\\
\qquad {\tt from} $\rightarrow$ Node[ $j=2$, {\tt from} := FromTableOne({\tt tab\_id}, {\tt condition}) ]\\
\qquad\qquad {\tt tab\_id} $\rightarrow$ Leaf[ $j=3$, {\tt tab\_id} := \emph{singer} ]\\
\qquad\qquad {\tt condition} $\rightarrow$ Node[ $j=4$, {\tt condition} := NoCondition() ]\\
\qquad {\tt condition} $\rightarrow$ Node[ $j=5$, {\tt condition} := NoCondition() ]\\
\qquad {\tt select} $\rightarrow$ Node[ $j=6$, {\tt select} := SelectColumnOne({\tt distinct}, {\tt col\_unit}) ]\\
\qquad\qquad {\tt distinct} $\rightarrow$ Node[ $j=7$, {\tt distinct} := False() ]\\
\qquad\qquad {\tt col\_unit} $\rightarrow$ Node[ $j=8$, {\tt col\_unit} := UnaryColumnUnit({\tt agg\_op}, {\tt distinct}, {\tt col\_id}) ]\\
\qquad\qquad\qquad {\tt distinct} $\rightarrow$ Node[ $j=9$, {\tt distinct} := False() ]\\
\qquad\qquad\qquad {\tt agg\_op} $\rightarrow$ Node[ $j=10$, {\tt agg\_op} := Count() ]\\
\qquad\qquad\qquad {\tt col\_id} $\rightarrow$ Leaf[ $j=11$, {\tt col\_id} := * ]\\
\qquad {\tt orderby} $\rightarrow$ Node[ $j=12$, {\tt orderby} := NoOrderBy() ]\\
    \hdashline
    \textbf{DB:} {\it flight\_2} \\
    \textbf{Question:} How many flights depart from City Aberdeen?\\
    \textbf{SQL:} {\tt select} {\tt count} ( * ) {\tt from} \emph{airports} {\tt join} {\it flights} {\tt on} \emph{flights.SourceAirport} = {\it airports.AirportCode} {\tt where} {\it airports.City} = ``{\it Aberdeen}'' \\
    \textbf{SQL AST:} \\
{\tt sql} $\rightarrow$ Node[ $j=0$, {\tt sql} := SQL({\tt from}, {\tt select}, {\tt condition}, {\tt groupby}, {\tt orderby}) ]\\
\qquad {\tt groupby} $\rightarrow$ Node[ $j=1$, {\tt groupby} := NoGroupBy() ]\\
\qquad {\tt condition} $\rightarrow$ Node[ $j=2$, {\tt condition} := CmpCondition({\tt col\_unit}, {\tt cmp\_op}, {\tt value}) ]\\
\qquad\qquad {\tt cmp\_op} $\rightarrow$ Node[ $j=3$, {\tt cmp\_op} := Equal() ]\\
\qquad\qquad {\tt col\_unit} $\rightarrow$ Node[ $j=4$, {\tt col\_unit} := UnaryColumnUnit({\tt agg\_op}, {\tt distinct}, {\tt col\_id}) ]\\
\qquad\qquad\qquad {\tt distinct} $\rightarrow$ Node[ $j=5$, {\tt distinct} := False() ]\\
\qquad\qquad\qquad {\tt agg\_op} $\rightarrow$ Node[ $j=6$, {\tt agg\_op} := None() ]\\
\qquad\qquad\qquad {\tt col\_id} $\rightarrow$ Leaf[ $j=7$, {\tt col\_id} := {\it airports.City} ]\\
\qquad\qquad {\tt value} $\rightarrow$ Node[ $j=8$, {\tt value} := LiteralValue({\tt val\_id}) ]\\
\qquad\qquad\qquad {\tt val\_id} $\rightarrow$ Leaf[ $j=9\textrm{-}10$, {\tt tok\_id} := ``{\it aberdeen <eos>}'' ]\\
\qquad {\tt select} $\rightarrow$ Node[ $j=11$, {\tt select} := SelectColumnOne({\tt distinct}, {\tt col\_unit}) ]\\
\qquad\qquad {\tt distinct} $\rightarrow$ Node[ $j=12$, {\tt distinct} := False() ]\\
\qquad\qquad {\tt col\_unit} $\rightarrow$ Node[ $j=13$, {\tt col\_unit} := UnaryColumnUnit({\tt agg\_op}, {\tt distinct}, {\tt col\_id}) ]\\
\qquad\qquad\qquad {\tt distinct} $\rightarrow$ Node[ $j=14$, {\tt distinct} := False() ]\\
\qquad\qquad\qquad {\tt agg\_op} $\rightarrow$ Node[ $j=15$, {\tt agg\_op} := Count() ]\\
\qquad\qquad\qquad {\tt col\_id} $\rightarrow$ Leaf[ $j=16$, {\tt col\_id} := \emph{*} ]\\
\qquad {\tt from} $\rightarrow$ Node[ $j=17$, {\tt from} := FromTableTwo({\tt tab\_id}, {\tt tab\_id}, {\tt condition}) ]\\
\qquad\qquad {\tt tab\_id} $\rightarrow$ Leaf[ $j=18$, {\tt tab\_id} := {\it airports} ]\\
\qquad\qquad {\tt tab\_id} $\rightarrow$ Leaf[ $j=19$, {\tt tab\_id} := {\it flights} ]\\
\qquad\qquad {\tt condition} $\rightarrow$ Node[ $j=20$, {\tt condition} := CmpCondition({\tt col\_unit}, {\tt cmp\_op}, {\tt value}) ]\\
\qquad\qquad\qquad {\tt cmp\_op} $\rightarrow$ Node[ $j=21$, {\tt cmp\_op} := Equal() ]\\
\qquad\qquad\qquad {\tt col\_unit} $\rightarrow$ Node[ $j=22$, {\tt col\_unit} := UnaryColumnUnit({\tt agg\_op}, {\tt distinct}, {\tt col\_id}) ]\\
\qquad\qquad\qquad\qquad {\tt distinct} $\rightarrow$ Node[ $j=23$, {\tt distinct} := False() ]\\
\qquad\qquad\qquad\qquad {\tt agg\_op} $\rightarrow$ Node[ $j=24$, {\tt agg\_op} := None() ]\\
\qquad\qquad\qquad\qquad {\tt col\_id} $\rightarrow$ Leaf[ $j=25$, {\tt col\_id} := \emph{flights.SourceAirport} ]\\
\qquad\qquad\qquad {\tt value} $\rightarrow$ Node[ $j=26$, {\tt value} := ColumnValue({\tt col\_id}) ]\\
\qquad\qquad\qquad\qquad {\tt col\_id} $\rightarrow$ Leaf[ $j=27$, {\tt col\_id} := \emph{airports.AirportCode} ]\\
\qquad {\tt orderby} $\rightarrow$ Node[ $j=28$, {\tt orderby} := NoOrderBy() ]\\
    \hdashline
    \textbf{DB:} {\it employee\_hire\_evaluation} \\
    \textbf{Question:} find the minimum and maximum number of products of all stores. \\
    \textbf{SQL:} {\tt select} {\tt max} ( {\it shop.Number\_products} )  , {\tt min} ( {\it shop.Number\_products} ) {\tt from} \emph{shop} \\
    \textbf{SQL AST:}\\
{\tt sql} $\rightarrow$ Node[ $j=0$, {\tt sql} := SQL({\tt from}, {\tt select}, {\tt condition}, {\tt groupby}, {\tt orderby}) ]\\
\qquad {\tt condition} $\rightarrow$ Node[ $j=1$, {\tt condition} := NoCondition() ]\\
\qquad {\tt from} $\rightarrow$ Node[ $j=2$, {\tt from} := FromTableOne({\tt tab\_id}, {\tt condition}) ]\\
\qquad\qquad {\tt tab\_id} $\rightarrow$ Leaf[ $j=3$, {\tt tab\_id} := {\it shop} ]\\
\qquad\qquad {\tt condition} $\rightarrow$ Node[ $j=4$, {\tt condition} := NoCondition() ]\\
\qquad {\tt groupby} $\rightarrow$ Node[ $j=5$, {\tt groupby} := NoGroupBy() ]\\
\qquad {\tt select} $\rightarrow$ Node[ $j=6$, {\tt select} := SelectColumnTwo({\tt distinct}, {\tt col\_unit}, {\tt col\_unit}) ]\\
\qquad\qquad {\tt distinct} $\rightarrow$ Node[ $j=7$, {\tt distinct} := False() ]\\
\qquad\qquad {\tt col\_unit} $\rightarrow$ Node[ $j=8$, {\tt col\_unit} := UnaryColumnUnit({\tt agg\_op}, {\tt distinct}, {\tt col\_id}) ]\\
\qquad\qquad\qquad {\tt distinct} $\rightarrow$ Node[ $j=9$, {\tt distinct} := False() ]\\
\qquad\qquad\qquad {\tt agg\_op} $\rightarrow$ Node[ $j=10$, {\tt agg\_op} := Max() ]\\
\qquad\qquad\qquad {\tt col\_id} $\rightarrow$ Leaf[ $j=11$, {\tt col\_id} := {\it shop.Number\_products} ]\\
\qquad\qquad {\tt col\_unit} $\rightarrow$ Node[ $j=12$, {\tt col\_unit} := UnaryColumnUnit({\tt agg\_op}, {\tt distinct}, {\tt col\_id}) ]\\
\qquad\qquad\qquad {\tt distinct} $\rightarrow$ Node[ $j=13$, {\tt distinct} := False() ]\\
\qquad\qquad\qquad {\tt agg\_op} $\rightarrow$ Node[ $j=14$, {\tt agg\_op} := Min() ]\\
\qquad\qquad\qquad {\tt col\_id} $\rightarrow$ Leaf[ $j=15$, {\tt col\_id} := {\it shop.Number\_products} ]\\
\qquad {\tt orderby} $\rightarrow$ Node[ $j=16$, {\tt orderby} := NoOrderBy() ]\\
    \hline

    \hline
    \end{tabular}%
    }
   \caption{Case study on traversal order. $j$ represents the decoding timestep when the node is expanded.}
  \label{tab:case_order1}%
\end{table*}%

\begin{table*}[htbp]
  \centering
  \resizebox{0.9\textwidth}{!}{
    \begin{tabular}{l}
    \hline

    \hline
\textbf{DB:} {\it cre\_Doc\_Template\_Mgt}\\
\textbf{Question:} What are the codes of template types that have fewer than 3 templates? \\
\textbf{SQL:} {\tt select} {\it template\_type\_code} {\tt from} \emph{Templates} {\tt group} {\tt by} {\it template\_type\_code} {\tt having} {\tt count} ( {\it *} ) < {\it 3} \\
\textbf{AST:} \\
{\tt sql} $\rightarrow$ Node[ $j=0$, {\tt sql} := SQL({\tt from}, {\tt select}, {\tt condition}, {\tt groupby}, {\tt orderby}) ]\\
\qquad {\tt condition} $\rightarrow$ Node[ $j=1$, {\tt condition} := NoCondition() ]\\
\qquad {\tt from} $\rightarrow$ Node[ $j=2$, {\tt from} := FromTableOne({\tt tab\_id}, {\tt condition}) ]\\
\qquad\qquad {\tt tab\_id} $\rightarrow$ Leaf[ $j=3$, {\tt tab\_id} := {\it Templates} ]\\
\qquad\qquad {\tt condition} $\rightarrow$ Node[ $j=4$, {\tt condition} := NoCondition() ]\\
\qquad {\tt groupby} $\rightarrow$ Node[ $j=5$, {\tt groupby} := GroupByColumnOne({\tt col\_id}, {\tt condition}) ]\\
\qquad\qquad {\tt col\_id} $\rightarrow$ Leaf[ $j=6$, {\tt col\_id} := {\it Templates.Template\_Type\_Code} ]\\
\qquad\qquad {\tt condition} $\rightarrow$ Node[ $j=7$, {\tt condition} := CmpCondition({\tt col\_unit}, {\tt cmp\_op}, {\tt value}) ]\\
\qquad\qquad\qquad {\tt cmp\_op} $\rightarrow$ Node[ $j=8$, {\tt cmp\_op} := LessThan() ]\\
\qquad\qquad\qquad {\tt col\_unit} $\rightarrow$ Node[ $j=9$, {\tt col\_unit} := UnaryColumnUnit({\tt agg\_op}, {\tt distinct}, {\tt col\_id}) ]\\
\qquad\qquad\qquad\qquad {\tt distinct} $\rightarrow$ Node[ $j=10$, {\tt distinct} := False() ]\\
\qquad\qquad\qquad\qquad {\tt agg\_op} $\rightarrow$ Node[ $j=11$, {\tt agg\_op} := Count() ]\\
\qquad\qquad\qquad\qquad {\tt col\_id} $\rightarrow$ Leaf[ $j=12$, {\tt col\_id} := \emph{*} ]\\
\qquad\qquad\qquad {\tt value} $\rightarrow$ Node[ $j=13$, {\tt value} := LiteralValue({\tt val\_id}) ]\\
\qquad\qquad\qquad\qquad {\tt val\_id} $\rightarrow$ Leaf[ $j=14\textrm{-}15$, {\tt tok\_id} := ``{\it 3 <eos>}'' ]\\
\qquad {\tt select} $\rightarrow$ Node[ $j=16$, {\tt select} := SelectColumnOne({\tt distinct}, {\tt col\_unit}) ]\\
\qquad\qquad {\tt distinct} $\rightarrow$ Node[ $j=17$, {\tt distinct} := False() ]\\
\qquad\qquad {\tt col\_unit} $\rightarrow$ Node[ $j=18$, {\tt col\_unit} := UnaryColumnUnit({\tt agg\_op}, {\tt distinct}, {\tt col\_id}) ]\\
\qquad\qquad\qquad {\tt distinct} $\rightarrow$ Node[ $j=19$, {\tt distinct} := False() ]\\
\qquad\qquad\qquad {\tt agg\_op} $\rightarrow$ Node[ $j=20$, {\tt agg\_op} := None() ]\\
\qquad\qquad\qquad {\tt col\_id} $\rightarrow$ Leaf[ $j=21$, {\tt col\_id} := \emph{Templates.Template\_Type\_Code} ]\\
\qquad {\tt orderby} $\rightarrow$ Node[ $j=22$, {\tt orderby} := NoOrderBy() ]\\
    \hdashline
\textbf{DB:} {\it student\_transcripts\_tracking}\\
\textbf{Question:} What is the first, middle, and last name of the first student to register?\\
\textbf{SQL:} {\tt select} {\it first\_name} ,  {\it middle\_name} ,  {\it last\_name} {\tt from} {\it Students} {\tt order by} {\it date\_first\_registered} {\tt asc} {\tt limit} 1\\
\textbf{AST:}\\
{\tt sql} $\rightarrow$ Node[ $j=0$, {\tt sql} := SQL({\tt from}, {\tt select}, {\tt condition}, {\tt groupby}, {\tt orderby}) ]\\
\qquad {\tt condition} $\rightarrow$ Node[ $j=1$, {\tt condition} := NoCondition() ]\\
\qquad {\tt from} $\rightarrow$ Node[ $j=2$, {\tt from} := FromTableOne({\tt tab\_id}, {\tt condition}) ]\\
\qquad\qquad {\tt tab\_id} $\rightarrow$ Leaf[ $j=3$, {\tt tab\_id} := {\it Students} ]\\
\qquad\qquad {\tt condition} $\rightarrow$ Node[ $j=4$, {\tt condition} := NoCondition() ]\\
\qquad {\tt groupby} $\rightarrow$ Node[ $j=5$, {\tt groupby} := NoGroupBy() ]\\
\qquad {\tt select} $\rightarrow$ Node[ $j=6$, {\tt select} := SelectColumnThree({\tt distinct}, {\tt col\_unit}, {\tt col\_unit}, {\tt col\_unit}) ]\\
\qquad\qquad {\tt distinct} $\rightarrow$ Node[ $j=7$, {\tt distinct} := False() ]\\
\qquad\qquad {\tt col\_unit} $\rightarrow$ Node[ $j=8$, {\tt col\_unit} := UnaryColumnUnit({\tt agg\_op}, {\tt distinct}, {\tt col\_id}) ]\\
\qquad\qquad\qquad {\tt distinct} $\rightarrow$ Node[ $j=9$, {\tt distinct} := False() ]\\
\qquad\qquad\qquad {\tt agg\_op} $\rightarrow$ Node[ $j=10$, {\tt agg\_op} := None() ]\\
\qquad\qquad\qquad {\tt col\_id} $\rightarrow$ Leaf[ $j=11$, {\tt col\_id} := \emph{Students.middle\_name} ]\\
\qquad\qquad {\tt col\_unit} $\rightarrow$ Node[ $j=12$, {\tt col\_unit} := UnaryColumnUnit({\tt agg\_op}, {\tt distinct}, {\tt col\_id}) ]\\
\qquad\qquad\qquad {\tt distinct} $\rightarrow$ Node[ $j=13$, {\tt distinct} := False() ]\\
\qquad\qquad\qquad {\tt agg\_op} $\rightarrow$ Node[ $j=14$, {\tt agg\_op} := None() ]\\
\qquad\qquad\qquad {\tt col\_id} $\rightarrow$ Leaf[ $j=15$, {\tt col\_id} := {\it Students.last\_name} ]\\
\qquad\qquad {\tt col\_unit} $\rightarrow$ Node[ $j=16$, {\tt col\_unit} := UnaryColumnUnit({\tt agg\_op}, {\tt distinct}, {\tt col\_id}) ]\\
\qquad\qquad\qquad {\tt distinct} $\rightarrow$ Node[ $j=17$, {\tt distinct} := False() ]\\
\qquad\qquad\qquad {\tt agg\_op} $\rightarrow$ Node[ $j=18$, {\tt agg\_op} := None() ]\\
\qquad\qquad\qquad {\tt col\_id} $\rightarrow$ Leaf[ $j=19$, {\tt col\_id} := {\it Students.first\_name} ]\\
\qquad {\tt orderby} $\rightarrow$ Node[ $j=20$, {\tt orderby} := OrderByLimitColumnOne({\tt col\_unit}, {\tt order}, {\tt val\_id}) ]\\
\qquad\qquad {\tt col\_unit} $\rightarrow$ Node[ $j=21$, {\tt col\_unit} := UnaryColumnUnit({\tt agg\_op}, {\tt distinct}, {\tt col\_id}) ]\\
\qquad\qquad\qquad {\tt distinct} $\rightarrow$ Node[ $j=22$, {\tt distinct} := False() ]\\
\qquad\qquad\qquad {\tt agg\_op} $\rightarrow$ Node[ $j=23$, {\tt agg\_op} := None() ]\\
\qquad\qquad\qquad {\tt col\_id} $\rightarrow$ Leaf[ $j=24$, {\tt col\_id} := \emph{Students.date\_first\_registered} ]\\
\qquad\qquad {\tt order} $\rightarrow$ Node[ $j=25$, {\tt order} := Asc() ]\\
\qquad\qquad {\tt val\_id} $\rightarrow$ Leaf[ $j=26\textrm{-}27$, {\tt tok\_id} := ``{\it 1 <eos>}'' ]\\
    \hline

    \hline
    \end{tabular}%
    }
   \caption{Case study on traversal order. $j$ represents the decoding timestep when the node is expanded.}
  \label{tab:case_order2}%
\end{table*}%

\section{Case Study on Decoder Self-attention}
\begin{figure}[ht]
\centering
  \begin{subfigure}{0.45\textwidth}
    \centering
    \includegraphics[width=\textwidth]{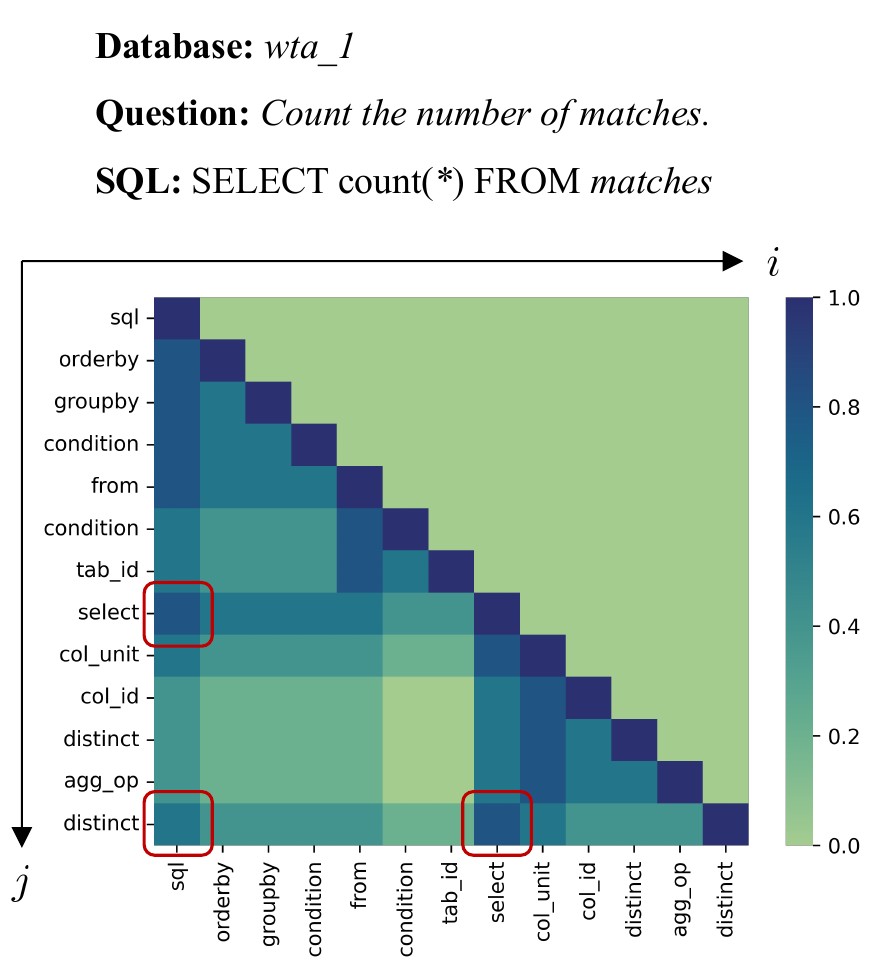}
    \caption{Heatmap matrix of golden labels calculated by Eq.(\ref{eq:golden}).}
    \label{fig:subfig1}
  \end{subfigure}
  \begin{subfigure}{0.45\textwidth}
    \centering
    \includegraphics[width=\textwidth]{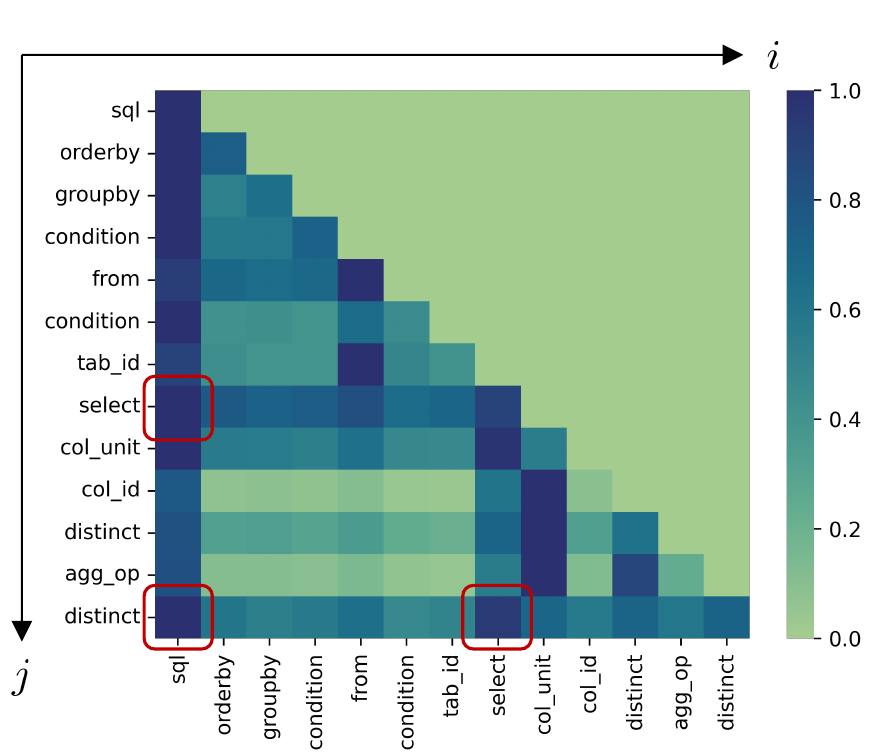}
    \caption{Heatmap matrix of model predictions.}
    \label{fig:subfig2}
  \end{subfigure}
  \caption{Visualization of target-side self-attention matrix among AST nodes.}
  \label{fig:subfigures}
\end{figure}
In this part, we visualize the heatmap matrix of target-side self-attention in Figure~\ref{fig:subfigures}. Assume that the relation between node $n_j$ and $n_i$ is $z_{ji}=(s, k)$, each entry with position $(j,i)$ in the golden heatmap matrix is computed via
\begin{align}
\text{score}(j,i)=1-\frac{s+k}{2M},\label{eq:golden}
\end{align}
where $M$ denotes the maximum depth of the current AST. In this way, we get an approximate scalar measure of the relative distance between any node pair. As for model predictions, we retrieve and take average of attention matrices $\alpha_{ji}^h$ from different heads $h$ in the last decoder layer. By comparison, we can find that: 1) The predicted attention matrix is highly similar to the golden reference based on path length. 2) Although some child/grandchild nodes are expanded in the distant future, they are still more correlated with their parent/grandparent, e.g., {\tt sql}$\rightarrow${\tt select}$\rightarrow${\tt distinct} in Figure~\ref{fig:subfigures}.
\end{document}